\documentclass[letterpaper]{article} 
\usepackage{aaai23}  
\usepackage{times}  
\usepackage{helvet}  
\usepackage{courier}  
\usepackage[hyphens]{url}  
\usepackage{graphicx} 
\usepackage{natbib}  
\usepackage{caption} 
\usepackage{algorithm}
\usepackage{algorithmic}
\usepackage{amssymb}
\usepackage{amsthm} 
\usepackage{amsmath}
\usepackage{amsfonts}
\usepackage{mathrsfs}
\usepackage{multirow}
\usepackage{makecell}
\usepackage{tablefootnote}
\usepackage{threeparttable}

\newtheorem{theorem}{Theorem}
\newtheorem{corollary}{Corollary}
\newtheorem{lemma}{Lemma}
\newtheorem{definition}{Definition}
\newtheorem{remark}{Remark}

\newenvironment{lemma1proof}{\noindent{\textbf{Proof of Lemma 1}:}}{\quad \hfill$\Box$}
\newenvironment{theorem1proof}{\noindent{\textbf{Proof of Theorem 1}:}}{\quad \hfill$\Box$}
\newenvironment{lemma2proof}{\noindent{\textbf{Proof of Lemma 2}:}}{\quad \hfill$\Box$}
\newenvironment{theorem2proof}{\noindent{\textbf{Proof of Theorem 2}:}}{\quad \hfill$\Box$}
\newenvironment{lemma3proof}{\noindent{\textbf{Proof of Lemma 3}:}}{\quad \hfill$\Box$}
\newenvironment{corollary1proof}{\noindent{\textbf{Proof of Corollary 1}:}}{\quad \hfill$\Box$}
\newenvironment{lemma4proof}{\noindent{\textbf{Proof of Lemma 4}:}}{\quad \hfill$\Box$}
\newenvironment{lemma5proof}{\noindent{\textbf{Proof of Lemma 5}:}}{\quad \hfill$\Box$}
\newenvironment{theorem3proof}{\noindent{\textbf{Proof of Theorem 3}:}}{\quad \hfill$\Box$}
\newenvironment{lemma6proof}{\noindent{\textbf{Proof of Lemma 6}:}}{\quad \hfill$\Box$}
\newenvironment{theorem4proof}{\noindent{\textbf{Proof of Theorem 4}:}}{\quad \hfill$\Box$}
\newenvironment{corollary2proof}{\noindent{\textbf{Proof of Corollary 2}:}}{\quad \hfill$\Box$}
\newenvironment{corollary3proof}{\noindent{\textbf{Proof of Corollary 3}:}}{\quad \hfill$\Box$}
\newenvironment{remark6proof}{\noindent{\textbf{Supplement to Remark 6}:}}{\quad \hfill$\Box$}
\newenvironment{remark5proof}{\noindent{\textbf{Supplement to Remark 5}:}}{\quad \hfill$\Box$}

\usepackage{newfloat}
\usepackage{listings}
\DeclareCaptionStyle{ruled}{labelfont=normalfont,labelsep=colon,strut=off} 
\floatstyle{ruled}
\newfloat{listing}{tb}{lst}{}
\floatname{listing}{Listing}
\title{On the Stability and Generalization of Triplet Learning}
\author{
    Jun Chen\textsuperscript{\rm 1},
    Hong Chen\textsuperscript{\rm 2, 3, 4,}\thanks{Corresponding author.}, Xue Jiang\textsuperscript{\rm 5}, Bin Gu\textsuperscript{\rm 6}, Weifu Li\textsuperscript{\rm 2, 3, 4}, Tieliang Gong\textsuperscript{\rm 7, 8}, Feng Zheng\textsuperscript{\rm 5}\\
}
\affiliations{
    \textsuperscript{\rm 1}College of Informatics, Huazhong Agricultural University, Wuhan 430070, China\\
    \textsuperscript{\rm 2}College of Science, Huazhong Agricultural University, Wuhan 430070, China\\
    \textsuperscript{\rm 3}Engineering Research Center of Intelligent Technology for Agriculture, Ministry of Education, Wuhan 430070, China\\
    \textsuperscript{\rm 4}Key Laboratory of Smart Farming for Agricultural Animals, Wuhan 430070, China\\
     \textsuperscript{\rm 5}Department of Computer Science and Engineering, Southern University of Science and Technology, Shenzhen 518055, China\\
     \textsuperscript{\rm 6}Mohamed bin Zayed University of Artificial Intelligence, Abu Dhabi, United Arab Emirates\\
    \textsuperscript{\rm 7}School of Computer Science and Technology, Xi'an Jiaotong University, Xi'an 710049, China\\
     \textsuperscript{\rm 8}Shaanxi Provincial Key Laboratory of Big Data Knowledge Engineering, Ministry of Education, Xi'an 710049, China\\
       chenh@mail.hzau.edu.cn
}

\usepackage{bibentry}

\begin{document}

\maketitle

\begin{abstract}
Triplet learning, i.e. learning from triplet data, has attracted much attention in computer vision tasks with an extremely large number of categories, e.g., face recognition and person re-identification. 
Albeit with rapid progress in designing and applying triplet learning algorithms, there is a lacking study on the theoretical understanding of their generalization performance. 
To fill this gap, this paper investigates the generalization guarantees of triplet learning by leveraging the stability analysis. 
Specifically, we establish the first general high-probability generalization bound for the triplet learning algorithm satisfying the uniform stability, and then obtain the excess risk bounds of the order $O(n^{-\frac{1}{2}} \mathrm{log}n)$ for both stochastic gradient descent (SGD) and regularized risk minimization (RRM), where $2n$ is approximately equal to the number of training samples. 
Moreover, an optimistic generalization bound in expectation as fast as $O(n^{-1})$ is derived for RRM in a low noise case via the on-average stability analysis. 
Finally, our results are applied to triplet metric learning to characterize its theoretical underpinning. 
\end{abstract}

\section{Introduction}
As two popular paradigms of machine learning, data-driven algorithms with pointwise loss and pairwise loss have been widely used to find the intrinsic relations from empirical observations.
In the algorithmic implementation, the former (called pointwise learning) often aims to minimize the empirical risk characterized by the divergence between the predicted output and the observed response of each input \cite{Vapnik1998, AMS/CuckerS2001, Nature/PoggioRMN2004}, while the latter (called pairwise learning) usually concerns the model performance associated with pairs of training instances, see e.g., ranking \cite{DBLP:journals/jmlr/AgarwalN09} and metric learning \cite{DBLP:conf/nips/XingNJR02, DBLP:journals/jmlr/YingL12}. 

Despite enjoying the advantages of feasible implementations and solid foundations, pointwise learning and pairwise learning may face a crucial challenge for learning tasks with an extremely large number of categories. 
Such learning scenarios appear in face recognition \cite{DBLP:conf/cvpr/SchroffKP15, DBLP:journals/pami/DingT18}, person re-identification \cite{DBLP:conf/nips/UstinovaL16, DBLP:conf/cvpr/ChengGZWZ16, DBLP:conf/cvpr/XiaoLOW16}, image retrieval \cite{DBLP:conf/cvpr/LaiPLY15, DBLP:conf/iccv/HuangFCY15} and other individual level fine-grained tasks \cite{DBLP:conf/cvpr/WohlhartL15, DBLP:conf/iccv/Simo-SerraTFKFM15}.
As illustrated in \citet{DBLP:conf/eccv/YuLGDT18}, the traditional learning model is difficult to achieve good performance in the setting of an extremely large number of categories since its parameters will increase linearly with the number of categories. 
To surmount this barrier, many triplet learning algorithms are formulated by injecting triplet loss function into the metric learning framework \cite{DBLP:conf/cvpr/SchroffKP15, DBLP:conf/nips/UstinovaL16, DBLP:conf/cvpr/ChengGZWZ16, DBLP:conf/cvpr/XiaoLOW16, DBLP:journals/pami/DingT18}. 
For triplet metric learning \cite{DBLP:conf/cvpr/SchroffKP15, DBLP:conf/eccv/GeHDS18}, the implementation procedures mainly include: 
1) Constructing triplets associated with anchor sample, positive sample and negative sample; 
2) Designing margin-based empirical risk associated with triplet loss; 
3) Learning metric space transformation rule via empirical risk minimization (ERM), which aims to minimize intra-class distance and maximize inter-class distance simultaneously. 
However, the triplet characteristic often leads to a heavy computational burden for large-scale data. 
Recently, stochastic gradient descent (SGD) is employed for deploying triplet learning algorithms due to its low time complexity 
  \cite{DBLP:conf/cvpr/SchroffKP15, DBLP:conf/eccv/GeHDS18}. 
Although there has been significant progress in designing and applying triplet learning algorithms, little work has been done to recover their generalization guarantees from the lens of statistical learning theory (SLT) \cite{Vapnik1998}. 

The generalization guarantee of learning algorithm is the core of SLT, which evaluates the prediction ability in the unseen inputs \cite{Vapnik1998, cucker_zhou_2007}. 
In a nutshell, there are three branches of generalization analysis including \emph{uniform convergence approaches} associated with hypothesis space capacity (e.g., VC dimension \cite{Vapnik1998}, covering numbers \cite{cucker_zhou_2007, DBLP:conf/nips/ChenWDH17}, Rademacher complexity \cite{DBLP:conf/colt/BartlettM01}), \emph{operator approximation technique} \cite{SmaleZ2007, DBLP:journals/jmlr/RosascoBV10}, and \emph{algorithmic stability analysis} \cite{DBLP:journals/jmlr/BousquetE02, DBLP:journals/jmlr/ElisseeffEP05, DBLP:journals/jmlr/Shalev-ShwartzSSS10}. 
It is well known that the stability analysis enjoys nice properties on flexibility (independent of the capacity of hypothesis function space) and adaptivity (suiting for rich learning scenarios, e.g., classification and regression \cite{DBLP:conf/icml/HardtRS16}, ranking \cite{DBLP:journals/jmlr/AgarwalN09}, and adversarial training \cite{DBLP:conf/nips/XingSC21}). 
Recently, besides learning algorithms based on ERM and regularized risk minimization (RRM), generalization and stability have been understood for SGD of pointwise learning \cite{DBLP:conf/icml/HardtRS16, DBLP:conf/nips/RouxSB12, DBLP:journals/jmlr/FehrmanGJ20, DBLP:journals/jmlr/LeiHT21} and pairwise learning \cite{DBLP:conf/nips/LeiLK20, DBLP:journals/jmlr/ArousGJ21, DBLP:conf/nips/LeiLY21}. While the existing extensive works on stability analysis, to our best knowledge, there is no related result of SGD and RRM for triplet learning. 

To fill the above gap, this paper aims to provide stability-based generalization analysis for a variety of triplet learning algorithms. 
We establish generalization bounds for SGD and RRM with triplet loss, which yield comparable convergence rates as pointwise learning \cite{DBLP:conf/colt/FeldmanV19} and pairwise learning \cite{DBLP:conf/nips/LeiLK20} under mild conditions.
The main contributions of this paper are summarized as follows.

\begin{itemize}
    \item \emph{Generalization by algorithmic stability for triplet learning}.
    After introducing a new definition of triplet uniform stability, we establish the first general high-probability generalization bound for triplet learning algorithms satisfying uniform stability, motivated by the recent analysis for pairwise learning \cite{DBLP:conf/nips/LeiLK20}. 
    Especially, the current analysis just requires the uniform stability of the triplet learning algorithm and the boundedness of loss function in expectation. 
    
    \item \emph{Generalization bounds for triplet SGD and triplet RRM}.
    Generalization properties are characterized for SGD and RRM of triplet learning when the loss function is (strongly) convex, $L$-Lipschitz and $\alpha$-smooth. 
    Particularly, the derived excess risk bounds are with the decay rate $O(n^{-\frac{1}{2}} \mathrm{log}n)$ as $n_+ \asymp n_-\asymp n$, where $n_+$ and $n_-$ are the numbers of positive samples and negative samples, respectively. 
    Moreover, for the strongly convex loss function, the refined generalization bound with the order $O(n^{-1})$ is derived for RRM by leveraging the triplet on-average stability. 
    To the best of our knowledge, these results are the first generalization bounds of SGD and RRM for triplet learning. 
\end{itemize}

\section{Related Work}
In this section, we briefly review the related works on triplet learning and algorithmic stability.

{\bf{Triplet learning.}}
The main purpose of deep metric learning is to directly learn a feature representation vector from input data with the help of deep neural networks. 
\citet{DBLP:conf/nips/BromleyGLSS93} found that the relationship between samples can be measured by the difference between the corresponding embedded vectors, and some deep metric learning models have been subsequently proposed \cite{DBLP:conf/cvpr/ChopraHL05, DBLP:conf/cvpr/HadsellCL06}. 
Later, \citet{DBLP:conf/cvpr/SchroffKP15} proposed the FaceNet by integrating the idea of triplet learning \cite{DBLP:conf/nips/SchultzJ03, DBLP:conf/nips/WeinbergerBS05} and deep metric learning together. 
In contrast to the previous approaches, FaceNet directly trains its output to be a compact 128-D embedding vector using a triplet loss function based on large margin nearest neighbor \cite{DBLP:conf/nips/WeinbergerBS05}, and it is implemented by employing the SGD strategy. 
Encouraged by the impressive performance of FaceNet, lots of learning algorithms with triplet loss have been formulated in the computer version field \cite{ DBLP:conf/cvpr/ChengGZWZ16, DBLP:conf/cvpr/XiaoLOW16, DBLP:conf/nips/UstinovaL16, DBLP:conf/cvpr/LiuTWPH16, DBLP:conf/cvpr/RamanathanLDHLG15,DBLP:journals/pami/DingT18}.
Although there have been significant works on designing triplet metric learning algorithms, our theoretical understanding of their generalization ability falls far below the experimental validations.

{\bf{Generalization and algorithmic stability.}}
In SLT, uniform convergence analysis focuses on bounding the uniform deviation between training error and testing error over hypothesis space \cite{Vapnik1998, AMS/CuckerS2001, DBLP:conf/colt/BartlettM01, DBLP:conf/nips/WangCZXGC20, DBLP:journals/tnn/ChenWZDH21}, and operator approximation approach is inspired by functional analysis theory \cite{SmaleZ2007, DBLP:journals/jmlr/RosascoBV10}. 
Indeed, the former depends on the capacity of hypothesis space (e.g., VC dimension \cite{Vapnik1998}, covering  numbers \cite{cucker_zhou_2007}, Rademacher complexity \cite{DBLP:conf/colt/BartlettM01}), and the latter is limited to some special models enjoying operator representation (e.g., regularized least squares regression \cite{SmaleZ2007}, regularized least squares ranking \cite{DBLP:journals/jat/Chen12}). 
Different from the above routes, algorithmic stability is described by the gap among training errors of different training sets, which is dimension-independent and enjoys adaptivity for wide learning models. 
The concept of algorithmic stability can be put forward as early as the 1970s \cite{Ann.Stat/RogersW1978}, and its learning theoretical framework was established in \citet{DBLP:journals/jmlr/BousquetE02} and \citet{DBLP:journals/jmlr/ElisseeffEP05}. 
In essential, the algorithmic uniform stability is closely related to the learnability \cite{Nature/PoggioRMN2004, DBLP:journals/jmlr/Shalev-ShwartzSSS10}. 
For the pointwise learning setting, the stability-based generalization guarantees have been stated in terms of uniform stability \cite{DBLP:conf/icml/HardtRS16, DBLP:conf/nips/FosterGKLMS19}, on-average stability \cite{DBLP:conf/icml/KuzborskijL18, DBLP:conf/iclr/LeiY21}, local elastic stability \cite{DBLP:conf/icml/DengHS21} and argument stability \cite{DBLP:conf/nips/BassilyFGT20, DBLP:conf/icml/LeiY20, DBLP:conf/icml/LiuLNT17}. 
For the pairwise learning setting, there are fine-grained analyses on the generalization and stability of SGD and RRM \cite{DBLP:journals/corr/ShenYYY19, DBLP:conf/nips/LeiLK20, DBLP:conf/nips/LeiLY21}. 
Due to the space limitation, we further summarize different definitions and properties of algorithmic stability in \emph{Supplementary Material C}. 
Along this line of the above corpus, it is natural to investigate the generalization bounds of triplet learning by algorithmic stability analysis. 

\section{Preliminaries}
This section introduces the necessary backgrounds on triplet learning and  algorithmic stability. The main notations used in this paper are stated in \emph{Supplementary Material A}.

\subsection{Triplet learning}
Let $\mathcal{X}_+, \mathcal{X}_- \subset \mathbb{R}^d$ are two $d$-dimensional input spaces and $\mathcal{Y} \subset \mathbb{R}$ is an output space.
We give the training set $S := \{z_{i}^{+} := (x_i^+, y_i^+)\}_{i=1}^{n_+} \cup \{z_{j}^{-} := (x_j^-, y_j^-)\}_{j=1}^{n_-}\in \mathcal{Z}$ with $\mathcal{Z}: = \mathcal{Z}_{+}^{n_{+}} \cup \mathcal{Z}_{-}^{n_{-}}$, where each positive sample $z_{i}^{+}$ and negative sample $z_{j}^{-}$ are drawn independently from $\mathcal{Z}_+ := \mathcal{X}_+ \times \mathcal{Y}$ and $\mathcal{Z}_- := \mathcal{X}_- \times \mathcal{Y}$, respectively.
Note that there are likely more than two classes in positive and negative sample spaces.
Given empirical observation $S$, triplet learning algorithms usually aim to find a model $h_w: \mathcal{X}_+ \times \mathcal{X}_+ \times \mathcal{X}_- \rightarrow \mathbb{R}$ such that the expectation risk 
\begin{eqnarray}
\label{expected risk}
    R(w) := \mathbb{E}_{z^+, \tilde{z}^+, z^-} \ell(w; z^+, \tilde{z}^+, z^-)
\end{eqnarray}
is as small as possible. Here the model parameter $w\in \mathcal{W}$ with $d^\prime$-dimensional parameter space $\mathcal{W} \subseteq \mathbb{R}^{d^{\prime}}$, $\mathbb{E}_z$ denotes the conditional expectation with respect to (w.r.t.) $z$, and the triplet loss function $\ell: \mathcal{W} \times \mathcal{Z}_+ \times \mathcal{Z}_+ \times \mathcal{Z}_- \rightarrow \mathbb{R}_+$ is used to measure the difference between model's prediction and corresponding real observation. 
Since the intrinsic distributions generating $z^+$ and $z^-$ are same and unknown, it is impossible to implement triplet learning by minimizing the objective $R(w)$ directly. 
Naturally, we consider the corresponding empirical risk of \eqref{expected risk} defined as
\begin{equation}
\label{empirical risk}
\begin{aligned}
    R_S(w) := \frac{1}{n_+(n_+-1)n_-} \sum\limits_{i,j\in [n_+], i \neq j, \atop k\in [n_-]} \ell(w; z_i^+, z_j^+, z_k^-)
\end{aligned}
\end{equation}
for algorithmic design, where $[n] := \{1, ..., n\}$.
Clearly, the triplet learning algorithms, built from $R_S(w)$ in \eqref{empirical risk}, are much more complicated than the corresponding ones in pointwise learning and pairwise learning. 

In the sequel, for the given algorithm $A$ and the training data $S$, we denote the corresponding output model parameter as $A(S)$ for feasibility. 
In triplet learning, we usually build predictors by optimizing the models measured by the empirical risk $R_S(A(S))$ or its variants. 
However, the nice empirical performance of learning model does not guarantee its effectiveness in unseen observations. 
In SLT, it is momentous and fundamental to bound generalization error, i.e. $R(w) - R_S(w)$, since it characterizes the gap between the population risk $R(w)$ and its empirical estimator $R_S(w)$. 
Despite the existing rich studies for pointwise learning and pairwise learning, the generalization bound of triplet learning is rarely touched in the machine learning community. 
In this paper, we pioneer the generalization analysis of triplet SGD and RRM to understand their learnability. 

\subsection{Triplet algorithmic stability}
An algorithm $A: \mathcal{Z}_+^{n_+} \cup \mathcal{Z}_-^{n_-} \rightarrow \mathcal{W}$ is stable if the model parameter $A(S)$ is insensitive to the slight change of training set $S$. 
Various definitions of algorithmic stability have been introduced from different motivations (see \emph{Supplementary Material C}), where uniform stability and on-average stability are popular for studying the generalization bounds of SGD and RRM \cite{DBLP:conf/icml/HardtRS16, DBLP:conf/icml/LinCR16, DBLP:conf/icml/KuzborskijL18, DBLP:conf/iclr/LeiY21, DBLP:conf/nips/LeiLK20}. 
Following this line, we extend the previous definitions of uniform stability and on-average stability to the triplet learning setting. 

\begin{definition}
\label{Definition of uniform stability}
(Uniform Stability). 
Assume any training datasets $S = \{z_1^+, ..., z_{n_+}^+, z_1^-, ..., z_{n_-}^-\}, \bar{S} = \{\bar{z}_1^+, ..., \bar{z}_{n_+}^+,\\ \bar{z}_1^-, ..., \bar{z}_{n_-}^-\}\in \mathcal{Z}_+^{n_+} \cup \mathcal{Z}_-^{n_-}$ are differ by at most a single sample. 
A deterministic algorithm $A: \mathcal{Z}_+^{n_+} \cup \mathcal{Z}_-^{n_-} \rightarrow \mathcal{W}$ is called $\gamma$-uniformly stable if 
\begin{equation*}
\begin{aligned}
    \mathop{\mathrm{sup}}\limits_{z^+, \tilde{z}^+\in \mathcal{Z}_+, \atop z^-\in \mathcal{Z}_-} |\ell(A(S); z^+, \tilde{z}^+, z^-)-
    \ell(A(\bar{S}); z^+, \tilde{z}^+, z^-)| \leq \gamma
\end{aligned}
\end{equation*}
for any training datasets $S, \bar{S} \in \mathcal{Z}_+^{n_+} \cup \mathcal{Z}_-^{n_-}$ that differ by at most a single sample.
\end{definition}
Definition \ref{Definition of uniform stability} coincides with the uniform stability definitions for pointwise learning \cite{DBLP:conf/icml/HardtRS16} and pairwise learning \cite{DBLP:conf/nips/LeiLK20}, except for the triplet loss involving two sample spaces $\mathcal{Z}_+$ and $\mathcal{Z}_-$. 

\begin{definition}
\label{Definition of on-average stability}
(On-average Stability). 
Let $S_{i,j,k} = \{z_1^+, ..., z_{i-1}^+, \bar{z}_{i}^+, z_{i+1}^+, ..., z_{j-1}^+, \bar{z}_{j}^+, z_{j+1}^+, ..., z_{n_+}^+, z_1^-, ...,\\ z_{k-1}^-, \bar{z}_{k}^-, z_{k+1}^-, ..., z_{n_-}^-\}, i,j \in [n_+], i\neq j, k\in [n_-]$. A deterministic algorithm $A: \mathcal{Z}_+^{n_+} \cup \mathcal{Z}_-^{n_-} \rightarrow \mathcal{W}$ is called  $\gamma$-on-average stable if 
\begin{equation*}
\begin{aligned}
    \frac{1}{n_+(n_+-1)n_-} \sum\limits_{i,j\in [n_+], i\neq j, \atop k\in [n_-]} & \mathbb{E}_{S, \bar{S}}\big[\ell(A(S_{i,j,k}); z_i^+, z_j^+, z_k^-)\\
    & - \ell(A(S); z_i^+, z_j^+, z_k^-)\big] \leq \gamma. 
\end{aligned}
\end{equation*}
\end{definition}
Compared with the existing ones for pointwise learning \cite{DBLP:conf/icml/KuzborskijL18} and pairwise learning \cite{DBLP:conf/nips/LeiLK20,DBLP:conf/nips/LeiLY21}, Definition \ref{Definition of on-average stability} considers much more complicated perturbations of training set $S$ involving three samples. 
Definition \ref{Definition of on-average stability} takes the expectation over $S$ and $\bar{S}$, and takes the average over perturbations, which is weaker than the uniform stability described in Definition \ref{Definition of uniform stability}.

\section{Main Results}
This section states our main results on generalization bounds for triplet learning by stability analysis. 
We show a general high-probability generalization bound of triplet learning algorithms firstly, and then apply it to two specific algorithms, i.e. SGD and RRM. 
Finally, the on-average stability is employed for getting an optimistic generalization bound of RRM in expectation. 
All the proofs are provided in \emph{Supplementary Material B} due to the space limitation. 

Similar with the previous analyses \cite{DBLP:conf/icml/HardtRS16, DBLP:conf/nips/LeiLK20}, our results are closely related to the following properties of triplet loss function.  

\begin{definition}
\label{Definition of some assumptions}
For a triplet loss function $\ell: \mathcal{W} \times \mathcal{Z}_+ \times \mathcal{Z}_+ \times \mathcal{Z}_- \rightarrow \mathbb{R}_+$, denote by $\nabla\ell(w):=\nabla\ell(w; z^+, \tilde{z}^+, z^-)$ its gradient w.r.t. the model parameter $w\in \mathcal{W}$ and denote by $\| \cdot \|$ a norm on an inner product space which satisfies $\| \cdot \|^2 = \langle \cdot, \cdot \rangle$. 
Let $\sigma \geq 0$ and $L, \alpha > 0$.

1) The triplet loss $\ell$ is $\sigma$-strongly convex if, for all $w, w^{\prime} \in \mathcal{W}$, 
\begin{equation*}
\begin{aligned}
    \ell(w) \geq \ell(w^{\prime}) + \langle \nabla\ell(w^\prime), w - w^\prime \rangle + \frac{\sigma}{2} \|w - w^\prime\|^2.
\end{aligned}
\end{equation*}

2) The triplet loss $\ell$ is L-Lipschitz if  
\begin{equation*}
\begin{aligned}
    |\ell(w) - \ell(w^\prime)| \leq L\|w - w^\prime\|, \forall w, w^{\prime} \in \mathcal{W}.
\end{aligned}
\end{equation*}

3) The triplet loss $\ell$ is $\alpha$-smooth if 
\begin{equation*}
\begin{aligned}
    \|\nabla \ell(w) - \nabla \ell(w^\prime)\| \leq \alpha \|w - w^\prime\|, \forall w, w^{\prime} \in \mathcal{W}.
\end{aligned}
\end{equation*}
\end{definition}

When $\sigma = 0$, $\ell$ is convex which also implies that $\nabla^2 \ell(w; z^+, \tilde{z}^+, z^-) > 0$. 
It is easy to verify that logistic loss, least square loss and Huber loss are convex and smooth. 
Meanwhile, we observe that hinge loss, logistic loss and Huber loss are convex and Lipschitz \cite{DBLP:conf/icml/HardtRS16, DBLP:conf/nips/LeiLK20, DBLP:conf/nips/LeiLY21}.

\subsection{Stability-based generalization bounds}
This subsection establishes the connection between uniform stability and generalization with high probability for triplet learning. 
Although rich results on the relationship between stability and generalization, the previous results do not hold directly for triplet learning due to its complicated loss structure. 
This difficulty is tackled by implementing much more detailed error decomposition and developing the analysis technique of \citet{DBLP:conf/nips/LeiLK20}. 
\begin{lemma}
\label{Lemma 1 of theorem 1}
If $A: \mathcal{Z}_+^{n_+} \cup \mathcal{Z}_-^{n_-} \rightarrow \mathcal{W}$ is $\gamma$-uniformly stable, for any $S, \bar{S}$, we have
\begin{equation*}
    |\ell(A(S); z^+, \tilde{z}^+, z^-) -\ell(A(S_{i,j,k}); z^+, \tilde{z}^+, z^-)| \leq 3\gamma
\end{equation*}
for all $z^+, \tilde{z}^+\in \mathcal{Z}_+, z^-\in \mathcal{Z}_-$, where $S, \bar{S}, S_{i, j, k}$ is defined in Definition \ref{Definition of on-average stability} for any $i,j \in [n_+], i \neq j, k\in [n_-]$.
\end{lemma}

Lemma \ref{Lemma 1 of theorem 1} illustrates that an upper bound of the change of the loss function still exists even after changing multiple samples of the training set. 
Here, the upper bound $3\gamma$ reflects the sensitivity of triplet learning w.r.t. the perturbation of training data. 

It is a position to state our first general generalization bound with high probability for the uniformly stable triplet learning algorithm $A$. 
Detailed proof can be found in \emph{Supplementary Material B.1}.

\begin{theorem}
\label{Theorem 1}
Assume  that $A: \mathcal{Z}_+^{n_+} \cup \mathcal{Z}_-^{n_-} \rightarrow \mathcal{W}$ is $\gamma$-uniformly stable. 
Let constant $M > 0$ and, for all $z^+, \tilde{z}^+ \in \mathcal{Z}_+$ and $ z^- \in \mathcal{Z}_-$, let $|\mathbb{E}_S\ell(A(S); z^+, \tilde{z}^+, z^-)| \leq M$.
Then, for all $\delta \in (0, 1/e)$, we have
\begin{eqnarray*}
    &&|R_S(A(S)) - R(A(S))|\\
    &\leq& 6\gamma + 
    e\Bigg(8M\Big(\frac{1}{\sqrt{n_-}} + \frac{2}{\sqrt{n_+ - 1}}\Big)\sqrt{\mathrm{log}(e/\delta)}\\
    &&+ 24\sqrt{2}\gamma \Big(\lceil \mathrm{log}_2(n_-(n_+ - 1)^2) \rceil + 2\Big) \mathrm{log}(e/\delta)\Bigg)
\end{eqnarray*}
with probability $1 - \delta$, where $\lceil n \rceil$ denotes the minimum integer no smaller than $n$ and $e$ denotes the base of the natural logarithm.
\end{theorem}

\begin{remark}
\label{Remark of theorem 1}
Theorem \ref{Theorem 1} demonstrates the generalization performance of triplet learning depends heavily on the sample numbers $n_+, n_-$ and the stability parameter $\gamma$, which extends the Theorem 1 of \citet{DBLP:conf/nips/LeiLK20} for pairwise learning to the triplet learning setting. 
Denote $x \asymp y$ as $ay < x \leq by$ for some constants $a, b > 0$. 
In particular, when $n_+ \asymp n_- \asymp n$, the high-probability bound in Theorem \ref{Theorem 1} can be rewritten as $O(n^{-\frac{1}{2}} + \gamma \mathrm{log}n)$, which is comparable with the previous analyses \cite{DBLP:conf/nips/LeiLK20, DBLP:conf/nips/LeiLY21}. 
\end{remark}
\begin{table*}[h]
    \begin{center}
        \renewcommand\arraystretch{1.5}
        \begin{tabular}{ccccccc}
            \hline
            \multirow{2}{*}{Algorithm} & \multirow{2}{*}{Reference} & \multicolumn{3}{c}{Assumptions} & \multirow{2}{*}{Tool} & \multirow{2}{*}{\shortstack{Convergence \\ rate}}\\
            & & Convex & Lipschitz & Smooth & &\\
            \hline
            \multirow{2}{*}{SGD ($\blacktriangle$)} & \citet{DBLP:conf/icml/HardtRS16} & $\surd$ & $\surd$ & $\surd$ & \thead{Uniform \\ stability} & $O(n^{-\frac{1}{2}})$\\
            & \citet{DBLP:conf/icml/LeiY20} & $\surd$ & $\times$ & $\surd$ & \thead{On-average \\ model stability} & $O(n^{-1})$\\
            \hline
            \multirow{5.5}{*}{SGD ($\blacktriangle\blacktriangle$)} & \citet{DBLP:conf/nips/LeiLK20} & $\surd$ & $\surd$ & $\surd$ & \thead{Uniform \\ stability} & $*O(n^{-\frac{1}{2}} \mathrm{log} n)$\\
            & \citet{DBLP:conf/nips/LeiLY21} & $\surd$ & $\times$ & $\surd$ & \thead{On-average \\ model stability} & $O(n^{-1})$\\
            & \citet{DBLP:conf/nips/LeiLY21} & $\surd$ & $\surd$ & $\times$ & \thead{On-average \\ model stability} & $O(n^{-\frac{1}{2}})$\\
            & \citet{DBLP:conf/nips/YangLWYY21} & $\surd$ & $\surd$ & $\times$ & \thead{Uniform \\ stability} & $O(n^{-\frac{1}{2}})$\\
            & \citet{DBLP:conf/nips/YangLWYY21} & $\surd$ & $\surd$ & $\surd$ & \thead{Uniform \\ stability} & $O(n^{-\frac{1}{2}})$\\
            \hline
            SGD ($\blacktriangle\blacktriangle\blacktriangledown$) & Ours ($n_+ \asymp n_- \asymp n$)& $\surd$ & $\surd$ & $\surd$ & \thead{Uniform \\ stability} & $*O(n^{-\frac{1}{2}} \mathrm{log} n)$\\
            \hline
        \end{tabular}
        \caption{Summary of stability-based generalization analyses of SGD in the setting of convexity ($\blacktriangle$-pointwise; $\blacktriangle\blacktriangle$-pairwise; $\blacktriangle\blacktriangle\blacktriangledown$-triplet; $\surd$-the reference has such a property; $\times$-the reference hasn't such a property; $*$-high-probability bound).}
        \label{Comparison1}
    \end{center}
\end{table*}

\subsection{Generalization bounds for SGD}
Let $w_1\in \mathcal{W}$ and let $\nabla \ell(w)$ be the subgradient of triplet loss $\ell$ w.r.t. the argument $w$. 
For triplet learning by SGD, at the $t$-th iteration, we draw $(i_t, j_t, k_t)$ randomly and uniformly over $\{(i_t, j_t, k_t): i_t, j_t\in [n_+], i_t \neq j_t, k_t\in [n_-]\}$, and update the model parameter $w_t$ by
\begin{equation}
\label{gradient update}
\begin{aligned}
    w_{t+1} = w_t - \eta_t \nabla\ell(w_t; z_{i_t}^+, z_{j_t}^+, z_{k_t}^-),
\end{aligned}
\end{equation}
where $\{\eta_t\}_t$ is a sequence of step sizes. 

To apply Theorem \ref{Theorem 1}, we need to bound the uniform stability parameter of \eqref{gradient update}.
Denote by $\mathbb{I}[\cdot]$ the indicator function which takes $1$ if the situation in the brackets is satisfied and takes $0$ otherwise.

\begin{lemma}
\label{Lemma 1 of theorem 2}
    Assume that $S, \bar{S} \in \mathcal{Z}_+^{n_+} \cup \mathcal{Z}_-^{n_-}$ are different only in the last positive sample (or negative sample).
    Suppose $\ell(w; z^+, \tilde{z}^+, z^-)$ is convex, $\alpha$-smooth and $L$-Lipschitz w.r.t. $\|\cdot\|, \forall z^+, \tilde{z}^+\in \mathcal{Z}_+, z^-\in \mathcal{Z}_-$. 
    If $\eta_t \leq 2/\alpha$, then SGD in \eqref{gradient update} with $t$-th iteration is $\gamma$-uniformly stable, where
    \begin{equation*}
    \begin{aligned}
        \gamma \leq & 2L^2 \sum\limits_{l=1}^{t} \eta_l \mathbb{I} \Big[(i_l = n_+ ~\mathrm{or}~ j_l = n_+, i_l \neq j_l, k_l \in [n_-], z_{n_+}^+\\ 
        & \neq \bar{z}_{n_+}^+) ~\mathrm{or}~ (i_l, j_l\in [n_+], i_l \neq j_l, k_l = n_-, z_{n_-}^-\neq \bar{z}_{n_-}^-)\Big].
    \end{aligned}
    \end{equation*}
\end{lemma}
In Lemma \ref{Lemma 1 of theorem 2}, we just consider the perturbation on the last positive (or negative) sample without loss of generality. 
The above uniform stability bound of SGD involves an indicator function associated with $S$ and $\bar{S}$, which is nonzero only when different triplets are used.

Now we state the generalization bounds for SGD \eqref{gradient update}. 
The proof is present in \emph{Supplementary Material B.2}.
\begin{theorem}
\label{Theorem 2}
    Let the loss function $\ell(w; z^+, \tilde{z}^+, z^-)$ is convex, $\alpha$-smooth and $L$-Lipschitz for all $z^+, \tilde{z}^+\in \mathcal{Z}_+, z^-\in \mathcal{Z}_-$ and  $|\mathbb{E}_S\ell(w_T; z^+, \tilde{z}^+, z^-)| \leq M$, where $w_T$ is produced by SGD \eqref{gradient update} with $\eta_t \equiv c/\sqrt{T}$ and constant $c\leq 2/\alpha$. 
    For any $\delta \in (0, 1/e)$, with probability $1 - \delta$ we have 
    \begin{equation*}
    \begin{aligned}
        &|R_S(w_T) - R(w_T)| \\
        = &O\Bigg(\Big(\lceil \mathrm{log}(n_-(n_+-1)^2)\rceil + 2 \Big) \mathrm{log}(1/\delta) \Big(\sqrt{\frac{\mathrm{log}(1/\delta)}{\mathrm{max}\{\frac{T}{n_+}, \frac{T}{n_-}\}}}\\
        &+ 1 \Big) \Big(\frac{\sqrt{T}}{n_+} + \frac{\sqrt{T}}{n_-} \Big) + \Big(\frac{1}{\sqrt{n_-}} + \frac{1}{\sqrt{n_+-1}} \Big) \sqrt{\mathrm{log}(1/\delta)} \Bigg).
    \end{aligned}
    \end{equation*}
\end{theorem}

\begin{remark}
\label{Remark of theorem 2}
Theorem \ref{Theorem 2} demonstrates that the generalization error of \eqref{gradient update} relies on the numbers of positive and negative training samples (i.e. $n_+, n_-$) and the iterative steps $T$. 
Our result also uncovers that the balance of positive and negative training samples is crucial to guarantee the generalization of triplet learning algorithms. 
When $n_+ \asymp n_- \asymp n$, we get the high-probability bound $|R_S(w_T) - R(w_T)| = O(n^{-\frac{1}{2}} \mathrm{log}n)$, which is consistent with Theorem 4 in \citet{DBLP:conf/nips/LeiLK20} for pariwise SGD.
\end{remark}

\begin{remark}
Let $w_R^* = \arg \min \limits_{w\in \mathcal{W}} R(w)$. 
We can deduce that 
\begin{eqnarray}
\label{excess risk decomposition}
    R(w_T) - R(w_R^*)=\big(R(w_T) - R_S(w_T)\big) + \big(R_S(w_T) \nonumber\\
    - R_S(w_R^*)\big) + \big(R_S(w_R^*) - R(w_R^*)\big).
\end{eqnarray}
As illustrated in previous studies \cite{DBLP:conf/nips/BottouB07, DBLP:conf/nips/LeiLK20, DBLP:conf/icml/LeiY20}, the first two terms in \eqref{excess risk decomposition} are called the estimation error and optimization error, respectively. 
Theorem \ref{Theorem 2} guarantees the upper bound of estimation error with $O(n^{-\frac{1}{2}}\mathrm{log}n)$ and \citet{DBLP:conf/colt/HarveyLPR19} states the upper  bound of the optimization error with $O(T^{-\frac{1}{2}} \mathrm{log}T)$. 
The third term on the right side of \eqref{excess risk decomposition} can be bounded by Bernstein's inequality for U-statistics \cite{ArXiv:Pitcan17}, which is present in the following Lemma \ref{Lemma 4 of theorem 2}.
\begin{lemma}
\label{Lemma 4 of theorem 2}
Let $b = \mathrm{sup}_{z^+, \tilde{z}^+, z^-} |\ell(w; z^+, \tilde{z}^+, z^-)|$ and $\tau$ be the variance of $\ell(w; z^+, \tilde{z}^+, z^-)$. 
Then, for any $\delta \in (0, 1)$, with probability at least $1 - \delta$ we have
\begin{equation*}
\begin{aligned}
    |R_S(w) - R(w)| \leq & \frac{2b\mathrm{log}(1/\delta)}{3\lfloor n_+/2 \rfloor} + \sqrt{ \frac{2\tau\mathrm{log}(1/\delta)}{\lfloor n_+/2 \rfloor}}\\
    & + \frac{2b\mathrm{log}(1/\delta)}{3\lfloor n_- \rfloor} + \sqrt{\frac{2\tau\mathrm{log}(1/\delta)}{\lfloor n_- \rfloor}},
\end{aligned}
\end{equation*}
where $\lfloor n \rfloor$ denotes the maximum integer no larger than $n$.
\end{lemma}
Under mild conditions, i.e., $b = O(\sqrt{n})$ and $n_+ \asymp n_- \asymp n$, we get $R_S(w_R^*) - R(w_R^*) = O\big(\frac{\mathrm{log}(1/\delta)}{\sqrt{n}} + \sqrt{\frac{\tau \mathrm{log} (1/\delta)}{n}}\big) = O(n^{-\frac{1}{2}})$. 
Combining this with the bounds of estimation error and optimization error in Remark 3, we deduce that the excess risk $R(w_T) - R(w_R^*) = O(n^{-\frac{1}{2}} \mathrm{log}n)$ as $T\asymp n$.
\end{remark}
\begin{remark}
To better highlight the characteristics of Theorem \ref{Theorem 2}, we compare it with the generalization analyses in the setting of convexity \cite{DBLP:conf/icml/HardtRS16, DBLP:conf/icml/LeiY20, DBLP:conf/nips/LeiLK20, DBLP:conf/nips/LeiLY21, DBLP:conf/nips/YangLWYY21} in Table \ref{Comparison1}. 
Clearly, our learning theory analysis is novel since it is the first touch for SGD under the triplet learning setting. 
When $n_+ \asymp n_- \asymp n$, the derived result is comparable with the previous convergence rates \cite{DBLP:conf/icml/HardtRS16, DBLP:conf/nips/LeiLK20, DBLP:conf/nips/LeiLY21}.
\end{remark}

\subsection{Generalization bounds for RRM}
We now turn to study the generalization properties of RRM for triplet learning. 
Detailed proofs are stated in \emph{Supplementary Material B.3}. 
Let $r: \mathcal{W}\rightarrow \mathbb{R}_{+}$ be a regularization penalty for increasing the data-fitting ability of ERM. 
For any datatset $S\in \mathcal{Z}_+^{n_+} \cup \mathcal{Z}_-^{n_-}$ and $R_S(w)$ defined in \eqref{empirical risk}, the derived model parameter of RRM is the minimizer of 
\begin{equation}
\label{RRM}
\begin{aligned}
    F_S(w):=R_S(w) + r(w)
\end{aligned}
\end{equation}
over $w\in \mathcal{W}$ and $F(w) := R(w) + r(w)$. 

To apply Theorem \ref{Theorem 1}, we also need to verify the stable parameter of RRM \eqref{RRM}.

\begin{lemma}
\label{Lemma 1 of theorem 3}
   Assume that $F_S(w)$ is $\sigma$-strongly convex w.r.t. $\|\cdot\|$ and $\ell(w; z^+, \tilde{z}^+, z^-)$ is convex and $L$-Lipschitz. 
   Then, the RRM algorithm $A$ defined as $A(S) = \arg \min \limits_{w\in \mathcal{W}} F_S(w)$ is $\gamma$-uniformly stable with $\gamma = \mathrm{min} \Big\{\frac{8}{n_+}, \frac{4}{n_-}\Big\} \frac{L^2}{\sigma}$.
\end{lemma}

When $n_+ \asymp n_- \asymp n$, the uniform stability parameter is $O(\frac{L^2}{n\sigma})$, which coincides with the previous analysis for pariwise learning \cite{DBLP:conf/nips/LeiLK20}. 
To tackle the triplet structure, the current analysis involves elaborate error decomposition and the deduce strategy of Lemma B.2 in \citet{DBLP:conf/nips/LeiLK20}. 

It is required in Theorem \ref{Theorem 1} that we assume the triplet loss for a uniformly stable algorithm is bounded in expectation. 
To get the necessary guarantee, we introduce the following Lemma \ref{Lemma 3 of theorem 3}, which can be proved coherently by utilizing Lemma 2 \cite{DBLP:conf/nips/LeiLK20} and the Lipschitz continuity of the loss function $\ell$. 

\begin{lemma}
\label{Lemma 3 of theorem 3}
    Let $F_S(w)$ be $\sigma$-strongly convex w.r.t. $\|\cdot\|$, $w^* = \arg \min \limits_{w\in\mathcal{W}} F(w)$, and, for all $z^+, \tilde{z}^+\in \mathcal{Z}_+, z^- \in \mathcal{Z}_-$, let 
    $\tilde{\ell}(A(S); z^+, \tilde{z}^+, z^-) = \ell(A(S); z^+, \tilde{z}^+, z^-) - \ell(w^*; z^+, \tilde{z}^+, z^-)$.
    If the RRM algorithm $A$ measured by loss function $\ell$ is $\gamma$-uniformly stable, then $A$ measured by loss function $\tilde{\ell}$ is also $\gamma$-uniformly stable and $$|\mathbb{E}_S\tilde{\ell}(A(S); z^+, \tilde{z}^+, z^-)| \leq M:= \mathrm{min}\Big\{ \frac{4\sqrt{6}}{\sqrt{n_+}}, \frac{4\sqrt{3}}{\sqrt{n_-}} \Big\}\frac{L^2}{\sigma}.$$
\end{lemma}

\begin{theorem}
\label{Theorem 3}
        Assume that $F_S(w)$ is $\sigma$-strongly convex w.r.t. $\|\cdot\|$, and $\ell(w; z^+, \tilde{z}^+, z^-)$ is convex and $L$-Lipschitz and $ \mathrm{sup}_{z^+, \tilde{z}^+, z^-} |\ell(w^*; z^+, \tilde{z}^+, z^-)| \leq O(\sqrt{n})$. 
        Let the variance of $\ell(w^*; z^+, \tilde{z}^+, z^-)$ is less than a positive constant $\tau$.
        For the RRM algorithm $A$ defined as $ $ and any $\delta\in (0, 1/e)$, we have
         \begin{equation*}
        \begin{aligned}
                & |R_S(A(S)) - R(A(S))|\\
                = & O\Bigg(\sigma^{-1} \Big( \mathrm{min}\Big\{\frac{\sqrt{2}}{\sqrt{n_+}}, \frac{1}{\sqrt{n_-}}\Big\} \Big(\frac{1}{\sqrt{n_-}} + \frac{1}{\sqrt{n_+}}\Big) \sqrt{\mathrm{log} \frac{1}{\delta}}\\
                & + \mathrm{min}\Big\{ \frac{2}{n_+}, \frac{1}{n_-} \Big\} \mathrm{log}\big(n_-n_+^2 \big) \mathrm{log} \frac1{\delta}\Big) + \sqrt{\frac{\mathrm{log}\frac1{\delta}}{n_+}}+ \sqrt{\frac{\mathrm{log}\frac1{\delta}}{n_-}} \Bigg)
             \end{aligned}
         \end{equation*}
        with probability $1 - \delta$.
\end{theorem}

\begin{remark}
\label{Remark of theorem 3}
If $n_+ \asymp n_- \asymp n$, the above bound is equivalent to $O\big(n^{-\frac{1}{2}} + (n\sigma)^{-1} \mathrm{log}n \big)$. Due to the definitions of $F_S(w)$ and $A(S)$ and $r(A(S))\geq 0$, we deduce that the excess risk $ R(A(S)) - R(w^*) \leq R(A(S)) - R_S(A(S)) + R_S(w^*)- R(w^*) +r(w^*)$.
Analogous to the third term to the right of \eqref{excess risk decomposition}, we have $R_S(w^*) - R(w^*) = O(\frac{\mathrm{log}(1/\delta)}{\sqrt{n}} + \sqrt{\frac{\tau \mathrm{log}(1/\delta)}{n}})$ with probability $1-\delta$. 
Therefore, the excess risk bound is $O(n^{-\frac{1}{2}} \mathrm{log}n)$ when $r(w^*) = O(\sigma \|w^*\|^2)$ and $\sigma \asymp n^{-\frac{1}{2}}$.
Note that the reason for $r(w^*) = O(\sigma \|w^*\|^2)$ can be found in the last part of Supplementary Material B.3.
\end{remark}

\subsection{Optimistic generalization bounds for RRM}
In this part, we use the on-average stability in Definition \ref{Definition of on-average stability} and some properties of smoothness to establish the optimistic generalization bounds of RRM in the low noise case. 
Different from the above theorems, we do not require the Lipschitz continuity condition for the triplet loss function.

The following lemma establishes the relationship between the estimation error and the model perturbation induced by the change at a single point of the training set.
\begin{lemma}
\label{Lemma 1 of theorem 4}
    Assume that for all $z^+, \tilde{z}^+\in \mathcal{Z}_+, z^-\in \mathcal{Z}_-$ and $w\in \mathcal{W}$, the loss function $\ell(w; z^+, \tilde{z}^+, z^-)$ is convex and $\alpha$-smooth w.r.t. $\|\cdot\|$.
    Then, for all $\epsilon > 0$,
    \begin{equation*}
    \begin{aligned}
        & \mathbb{E}_S[R(A(S)) - R_S(A(S))]\\
        \leq & \frac{3(\epsilon + \alpha)}{2n_+(n_+-1)n_-} \sum\limits_{i\in [n_+], \atop k\in [n_-]} \Big( 2\mathbb{E}_{S,\bar{S}}\|A(S_i) - A(S)\|^2\\
        & + \mathbb{E}_{S,\bar{S}}\|A(S_k) - A(S)\|^2 \Big) + \frac{\alpha \mathbb{E}_SR_S(A(S))}{\epsilon}, 
    \end{aligned}
    \end{equation*}
    where $S_i = \{z_1^+, ..., z_{i-1}^+, \bar{z}_i^+, z_{i+1}^+, ..., z_{n_+}^+, z_1^-, ..., z_{n_-}^-\}$ and $S_k = \{z_1^+, ..., z_{n_+}^+, z_1^-, ..., z_{k-1}^-, \bar{z}_k^-, z_{k+1}^-, ..., z_{n_-}^-\}$. 
\end{lemma}

From the proof of Lemma \ref{Lemma 1 of theorem 4} (see \emph{Supplementary Material B.4}) and Definition \ref{Definition of on-average stability},  we know the upper bound in Lemma  \ref{Lemma 1 of theorem 4} provides the selection of on-average stability parameter $\gamma$.
After establishing the connection between $\mathbb{E}_{S, \bar{S}}\|A(S_{i}) - A(S)\|^2$ (or $\mathbb{E}_{S, \bar{S}}\|A(S_{k}) - A(S)\|^2$) and $\mathbb{E}_SR_S(A(S))$, we get the following error bound of RRM. 

\begin{table*}[!h]
    \begin{center}
        \renewcommand\arraystretch{1.5}
        \begin{tabular}{ccccccccc}
            \hline
            \multirow{2.5}{*}{Algorithm} & \multirow{2.5}{*}{Reference} & \multicolumn{3}{c}{Assumptions} & \multirow{2.5}{*}{Tool} & \multirow{2.5}{*}{\shortstack{Convergence \\ rate}}\\
            & & \thead{Strongly \\ Convex} & Lipschitz & Smooth & &\\
            \hline
            \thead{Full-batch \\ SGD ($\blacktriangle$)} & \citet{DBLP:conf/nips/KlochkovZ21} & $\surd$ & $\surd$ & $\times$ & \thead{Uniform \\  stability} & $*O(n^{-1} \mathrm{log} n)$\\
            RRM ($\blacktriangle$) & \citet{DBLP:conf/colt/FeldmanV19} & $\surd$ & $\surd$ & $\times$ & \thead{Uniform \\ stability} & $*O(n^{-\frac{1}{2}} \mathrm{log} n)$\\
            \hline
            \multirow{2.75}{*}{RRM ($\blacktriangle\blacktriangle$)} & \citet{DBLP:conf/nips/LeiLK20} & $\surd$ & $\surd$ & $\times$ & \thead{Uniform \\ stability} & $*O(n^{-\frac{1}{2}} \mathrm{log} n)$\\
            & \citet{DBLP:conf/nips/LeiLK20} & $\surd$ & $\times$ & $\surd$ & \thead{On-average \\ stability} & $O(n^{-1})$\\
            \hline
            \multirow{2.75}{*}{RRM ($\blacktriangle\blacktriangle\blacktriangledown$)} & Ours ($n_+ \asymp n_- \asymp n$) & $\surd$ & $\surd$ & $\times$ & \thead{Uniform \\ stability} & $*O(n^{-\frac{1}{2}} \mathrm{log} n)$\\
            & Ours ($n_+ \asymp n_- \asymp n$) & $\surd$ & $\times$ & $\surd$ & \thead{On-average \\ stability} & $O(n^{-1})$\\
            \hline
        \end{tabular}
        \caption{Summary of stability-based generalization analyses for algorithms in the setting of strong convexity ($\blacktriangle$-pointwise; $\blacktriangle\blacktriangle$-pairwise; $\blacktriangle\blacktriangle\blacktriangledown$-triplet; $\surd$-the reference has such a property; $\times$-the reference hasn't such a property; $*$-high-probability bound).}
        \label{Comparison2}
    \end{center}
\end{table*}

\begin{theorem}
\label{Theorem 4}
    Assume that the loss function $\ell(w; z^+, \tilde{z}^+, z^-)$ is convex and $\alpha$-smooth for all $z^+, \tilde{z}^+\in \mathcal{Z}_+, z^-\in \mathcal{Z}_-$ and $w\in \mathcal{W}$, and $F_S(w)$ is $\sigma$-strongly convex w.r.t. $\|\cdot\|$ with $S\in \mathcal{Z}_+^{n_+}\cup \mathcal{Z}_-^{n_-}$.
    Let $\sigma \mathrm{min}\{n_+,n_-\} \geq 8\alpha$ and $A(S) = \arg \min\limits_{w\in \mathcal{W}}F_S(w)$.
    Then, for all $\epsilon > 0$,
    \begin{equation*}
    \begin{aligned}
        & \mathbb{E}_S[F(A(S)) - F_S(w^*)]
        \leq  \mathbb{E}_S[R(A(S)) - R_S(A(S))]\\
        \leq & \bigg(\frac{\alpha}{\epsilon} + \frac{1536\alpha(\epsilon + \alpha)}{n_+^2(n_+-1)\sigma^2} + \frac{256\alpha(\epsilon + \alpha)}{3(n_+-1)n_-^2\sigma^2}\bigg) \mathbb{E}_SR_S(A(S)).
    \end{aligned}
    \end{equation*}
\end{theorem}

\begin{remark}
\label{Remark of theorem 4}
The upper bounds in Theorem \ref{Theorem 4} are closely related to the empirical risk $\mathbb{E}_SR_S(A(S))$.
It is reasonable to assume that the empirical risk of $A(S)$ is small enough with the increasing of training samples. 
When $n_+\asymp n_-\asymp n, r(w^*) = O(\sigma \|w^*\|^2)$ and $\epsilon = \sqrt{\frac{3 n_+^2 (n_+-1) n_-^2 \sigma^2}{4608 n_-^2 + 256 n_+^2}}$, $\mathbb{E}_S\big[R(A(S)) - R(w^*)\big] = O\bigg(\frac{ R(w^*)}{n^{\frac{3}{2}}\sigma}+ \big(n^{-\frac{3}{2}} + \sigma \big)\|w^*\|^2 \bigg)$.
When $\sigma = n^{-\frac{3}{4}}\|w^*\|^{-1}\sqrt{R(w^*)}$ and $R(w^*) = n^{-\frac{1}{2}} \|w^*\|^2$, $\mathbb{E}_S\big[R(A(S)) - R(w^*)\big] = O(n^{-1} \|w^*\|^2)$.
Note that $R(w^*)$ can not less than $n^{-\frac{1}{2}} \|w^*\|^2$ due to $\sigma \mathrm{min}\{n_+,n_-\} \geq 8\alpha$. 

The above excess risk bound assures the convergence rate  $O(n^{-1} \|w^*\|^2)$ in expectation under proper conditions of $w^*$ and $R_S(A(S))$, which extends the previous optimistic generalization bounds of pointwise learning \cite{DBLP:conf/nips/SrebroST10, DBLP:conf/colt/ZhangYJ17} and pairwise learning \cite{DBLP:conf/nips/LeiLK20} to the triplet setting. 
\end{remark}

\begin{remark}
As summarized in Table \ref{Comparison2}, the convergence guarantees in Theorems \ref{Theorem 3} and \ref{Theorem 4} are comparable with the existing results in the setting of strong convexity even involving the complicated triplet structure in error decomposition.  
\end{remark}

\section{Applied to Triplet Metric Learning}
This section applies our generalization analysis to triplet metric learning, which focuses on learning a metric to minimize the intra-class distance and maximize inter-class distance simultaneously. Let $t(y, y^\prime)$ be the symbolic function, i.e., $t(y, y^\prime) = 1$ if $y = y^\prime$ and $-1$ otherwise.
Inspired by the 0-1 loss in pairwise metric learning $\ell_{0-1}(w; z, z^{\prime}) = \mathbb{I}[t(y, y^{\prime}) (1-h_w(x, x^{\prime}))\leq 0]$ \cite{DBLP:conf/nips/LeiLY21}, we consider a 0-1 triplet loss $\ell_{0-1}(w; z^+, \tilde{z}^+, z^-) = \mathbb{I}[h_w(x^+, \tilde{x}^+) - h_w(x^+, x^-) + \zeta \geq 0]$, where the training model $h_w$ is considered as $h_w(x^+, \tilde{x}^+) = \big\langle w, (x^+ - \tilde{x}^+)(x^+ - \tilde{x}^+)^\top \big\rangle$, and $\zeta$ denotes the margin that requires the distance of negative pairs to excess the one of positive pairs. 
We introduce the triplet loss 
\begin{equation}
\label{metric-loss}
    \ell_{\phi}(w; z^+, \tilde{z}^+, z^-) = \phi (h_w(x^+, \tilde{x}^+) - h_w(x^+, x^-) + \zeta)
\end{equation}
associated with the logistic function $\phi(u) = \mathrm{log}(1 + \mathrm{exp}(-u))$, which is consistent with the error metric used in \citet{DBLP:conf/cvpr/SchroffKP15} and \citet{DBLP:conf/eccv/GeHDS18}. 

When $\max\{\mathrm{sup}_{x^+\in \mathcal{X}_+}\|x^+\|, \mathrm{sup}_{x^-\in \mathcal{X}_-}\|x^-\|\} \leq B$, Theorems \ref{Theorem 2}-\ref{Theorem 3} yield the following convergence rates for SGD and RRM with the triplet loss \eqref{metric-loss}, respectively.
\begin{corollary}
\label{Corollary 1}
Let $w_T$ is produced by SGD \eqref{gradient update} with $\eta_t \equiv c/\sqrt{T}, c\leq 1/(32B^4)$ and $\big|\mathbb{E}_S[\phi (h_{w_T}(x^+,\tilde{x}^+) - h_{w_T}(x^+, x^-) + \zeta)]\big| \leq M$. 
For any $\delta \in (0, 1/e)$, with probability $1 - \delta$, we have
$|R_S(w_T) - R(w_T)| = O\big(n^{-\frac{1}{2}} \mathrm{log}n \mathrm{log}^{\frac{3}{2}}(1/\delta) + n^{-\frac{1}{2}} \mathrm{log}^{\frac{1}{2}}(1/\delta) \big)$.
\end{corollary}

\begin{corollary}
\label{Corollary 2}
Consider $F_S(w)$ in \eqref{RRM} with the triplet loss \eqref{metric-loss} and $r(w^*)= O(\sigma \|w^*\|^2)$ with $\sigma \asymp n^{-\frac{1}{2}}$. 
Assume that $\mathrm{sup}_{z^+, \tilde{z}^+, z^-} |\ell(w^*; z^+, \tilde{z}^+, z^-)|\leq O(\sqrt{n})$ and the variance of $\ell(w^*; z^+, \tilde{z}^+, z^-) $ is bounded.
Then for $A(S) = \arg \min \limits_{w\in \mathcal{W}} F_S(w)$ and any $\delta \in (0, 1/e)$, we have $R(A(S)) - R(w^*) = O(n^{-\frac{1}{2}} \mathrm{log}n \mathrm{log}(1/\delta))$ with probability $1 - \delta$.
\end{corollary}
Moreover, we get the refined result of RRM from Theorem \ref{Theorem 4} with the help of the strong-convexity of \eqref{metric-loss}. 
\begin{corollary}
\label{Corollary 3}
    Under the basic assumptions and notations of Corollary \ref{Corollary 2}, assume $\sigma = n^{-\frac{3}{4}} \|w^*\|^{-1} \sqrt{R(w^*)}$ and $R(w^*) = n^{-\frac{1}{2}} \|w^*\|^2$, then we have $\mathbb{E}_S\big[R(A(S)) - R(w^*)\big] = O(n^{-1} \|w^*\|^2).$
\end{corollary}

\section{Conclusion}
This paper fills the theoretical gap in the generalization bounds of SGD and RRM for triplet learning by developing algorithmic stability analysis techniques, which are valuable to understanding their intrinsic statistical foundations of outstanding empirical performance. 
We firstly derive the general high-probability generalization bound $O(\gamma \mathrm{log} n + n^{-\frac{1}{2}})$ for triplet uniformly stable algorithms, and then apply it to get the explicit result $O(n^{-\frac{1}{2}} \mathrm{log} n)$ for SGD and RRM under mild conditions of loss function. 
For RRM with triplet loss, the optimistic bound  $O(n^{-1})$ in expectation is also provided by leveraging the on-average stability. 
Even for the complicated triplet structure, our results also enjoy similar convergence rates as the previous related works of pointwise learning \cite{DBLP:conf/icml/HardtRS16,DBLP:conf/colt/FeldmanV19} and pairwise learning \cite{DBLP:conf/nips/LeiLK20}. 
Some potential directions are discussed in \emph{Supplementary Material D} for future research.

\section{Acknowledgments}
This work was supported  in part by National
Natural Science Foundation of China under Grant Nos. 12071166, 62106191, 61972188, 62122035.

\bibliography{reference}{}

\newpage
\onecolumn

\setcounter{equation}{7}   
\setcounter{lemma}{6}   
\setcounter{table}{2}   

\section{A.~~ Notations}
The main notations of this paper are summarized in Table \ref{Notations}.
\begin{table}[!ht]
    \centering
    \renewcommand\arraystretch{1.5}
    \begin{tabular}{l|l}
        \hline
        Notations & Descriptions\\
        \hline
        SGD & Stochastic gradient descent\\
        RRM & Regularized risk minimization\\
        ERM & Empirical risk minimization\\
        SLT & Statistical learning theory\\
        $\mathcal{Z}_+ (\mathcal{Z}_-)$ & the compact positive (negative) sample space associated with input space $\mathcal{X}_+$ ($\mathcal{X}_-$ ) and output set  $\mathcal{Y}$\\
        $z^+ = (x^+, y^+)$ & the random sample sampling from $\mathcal{Z}_+$ \\
        $z^- = (x^-, y^-)$ & the random sample sampling from $\mathcal{Z}_-$\\
        $n_+,n_-$ & the numbers of samples sampling from $\mathcal{Z}_+$ and $\mathcal{Z}_-$, respectively\\
        $h_w$ & the training model\\
        $w, \mathcal{W}$ & the parameter of training model and model parameter space, respectively\\
        $S$ & the training dataset defined as $\{z_{i}^{+} = (x_i^+, y_i^+)\}_{i=1}^{n_+} \cup \{z_{j}^{-} = (x_j^-, y_j^-)\}_{j=1}^{n_-}\in \mathcal{Z}: =  \mathcal{Z}_{+}^{n_{+}} \cup \mathcal{Z}_{-}^{n_{-}}$\\
        $d, d^\prime$ & the dimensions of $\mathcal{X}_+ (\mathcal{X}_-)$ and $\mathcal{W}$, respectively\\
        $\ell(w)$ & the triplet loss function defined as $\ell(w; z^+, \tilde{z}^+, z^-)$\\
        $\nabla \ell$ & the  gradient of $\ell(w; z^+, \tilde{z}^+, z^-)$ to the first argument $w$\\
        $R, R_S$ & the population risk and empirical risk based on training dataset $S$, respectively\\
        $T$ & the number of iterative steps for SGD\\
        $w_T$ & the model parameter derived by SGD after  $T$-th update\\
        $\eta_T$ & the step size at the $T$-th update\\
        $A, A(S)$ & the given algorithm and its output model parameter based on training dataset $S$, respectively\\
        $\gamma, \sigma, L, \alpha$ & the parameters of stability, strong convexity, Lipschitz continuity and smoothness, respectively\\
        $\tau$& the variance of triplet loss $\ell(w; z^+, \tilde{z}^+, z^-)$\\
        $r(w)$ & the regularization term
        \\
        $F_S(w)$ & the regularized empirical risk defined as $R_S(w) + r(w)$\\
        $w_R^*$ &  the optimal model based on the expected risk, $w_R^* = \arg \min\limits_{w\in \mathcal{W}} R(w)$\\
        $w^*$ &  the optimal model based on the regularized empirical risk, $w^* = \arg \min\limits_{w\in \mathcal{W}} F_S(w)$\\
        $t(y, y^\prime)$ & the symbolic function, i.e., $t(y, y^\prime) = 1$ if $y = y^\prime$ and $-1$ otherwise\\ 
        $\phi$ & the logistic function $\phi(u) = \mathrm{log}(1 + \mathrm{exp}(-u))$ \\
        $\top$ & the transpose of a vector or a matrix\\
        $\asymp$ & $n_+ \asymp n_-$ if there exist positive constants $c_1,c_2$ such that $c_1n_+ \leq n_-\leq c_2n_+$\\
        $[\cdot]$ & $[n] := \{1, ..., n\}$\\
        $\lfloor \cdot \rfloor$, $\lceil \cdot \rceil$ & $\lfloor n \rfloor$: the maximum integer no larger than $n$, $\lceil n \rceil$: the minimum integer no smaller than $n$\\
        $e$ & the base of the natural logarithm\\
        $\mathbb{I}[\cdot]$ & the indicator function\\
        $\zeta$ & the margin that requires the distance of negative pairs to excess one of the positive pairs\\
        \hline
    \end{tabular}
    \caption{Summary of main notations involved in this paper.}
    \label{Notations}
\end{table}

\section{B. Proofs of Main Results}
\begin{figure}[h]
    \centering
    \includegraphics[scale=0.39]{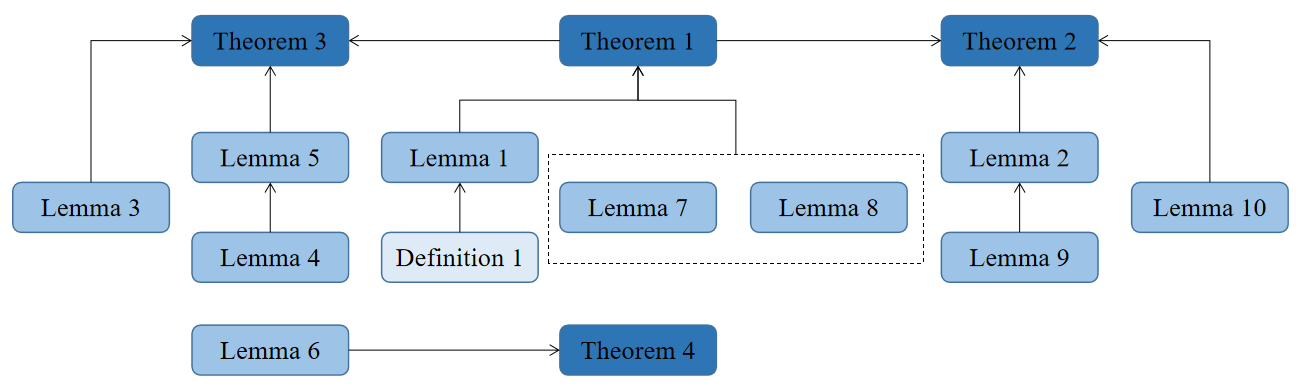}
    \caption{The proof diagram for Theorems 1-4.}
    \label{Proof sketch}
\end{figure}

We start with the sketching of the relations among theorems and lemmas in Figure \ref{Proof sketch}, and then progress to the detailed proofs.

Note that our framework is similar with \citet{DBLP:conf/nips/LeiLK20}, which mainly includes three parts: stability definition, the quantitative relationship between stability and generalization error, and the upper bound of stability parameter. Our theoretical results cannot be established directly from the current framework due to the complex triplet loss. To fill this theoretical gap on the generalization guarantees of triplet learning, we develop stability analysis technique by constructing new definitions of stability(see Definitions 1, 2), introducing detailed error decomposition(see Theorems 1), and considering fine-grained sampling situations for bounding stability parameter(see proofs of Lemmas 2, 4 and Theorems 2, 4).

   

\subsection{B.1~~~ Proof of Theorem 1}
Lemma 1 can be proved with the help of the triangular inequality and Definition 1.

\begin{lemma1proof}
Recall that, $S$ and $\bar{S}$ involved in Definition 1, differ by at most a single example. 
To relate $S$ with 
$$S_{i,j,k} = \{z_1^+, ..., z_{i-1}^+, \bar{z}_{i}^+, z_{i+1}^+, ..., z_{j-1}^+, \bar{z}_{j}^+, z_{j+1}^+, ..., z_{n_+}^+, z_1^-, ..., z_{k-1}^-, \bar{z}_{k}^-, z_{k+1}^-, ..., z_{n_-}^-\}, i,j \in [n_+], i\neq j, k\in [n_-]$$ (also see Definition 2), we introduce the following stepping-stone sets 
\begin{equation*}
\begin{aligned}
    S_i = \{z_1^+, ..., z_{i-1}^+, \bar{z}_i^+, z_{i+1}^+, ..., z_{n+}^+, z_1^-, ..., z_{n_-}^-\}
\end{aligned}
\end{equation*}
and
\begin{equation*}
\begin{aligned}
    S_{i,j} = \{z_1^+, ..., z_{i-1}^+, \bar{z}_i^+, z_{i+1}^+, ..., z_{j-1}^+, \bar{z}_j^+, z_{j+1}^+, ..., z_{n+}^+, z_1^-, ..., z_{n_-}^-\}, \forall i,j\in [n_+], i \neq j.
\end{aligned}
\end{equation*}
It is easy to verify that
\begin{equation*}
\begin{aligned}
    &\mathop{\mathrm{sup}}\limits_{z^+, \tilde{z}^+\in \mathcal{Z}_+, \atop z^-\in \mathcal{Z}_-} |\ell(A(S); z^+, \tilde{z}^+, z^-) - \ell(A(S_{i,j,k}); z^+, \tilde{z}^+, z^-)| \\
    \leq & \mathop{\mathrm{sup}}\limits_{z^+, \tilde{z}^+\in \mathcal{Z}_+, \atop z^-\in \mathcal{Z}_-} |\ell(A(S); z^+, \tilde{z}^+, z^-) - \ell(A(S_{i}); z^+, \tilde{z}^+, z^-)| 
    + \mathop{\mathrm{sup}}\limits_{z^+, \tilde{z}^+\in \mathcal{Z}_+, \atop z^-\in \mathcal{Z}_-} |\ell(A(S_i); z^+, \tilde{z}^+, z^-) - \ell(A(S_{i, j}); z^+, \tilde{z}^+, z^-)| \\    
    & + \mathop{\mathrm{sup}}\limits_{z^+, \tilde{z}^+\in \mathcal{Z}_+, \atop z^-\in \mathcal{Z}_-} |\ell(A(S_{i,j}); z^+, \tilde{z}^+, z^-) - \ell(A(S_{i,j,k}); z^+, \tilde{z}^+, z^-)| \\
    \leq & 3 \gamma.
\end{aligned}
\end{equation*}

This completes the proof.
\end{lemma1proof}

Now we introduce the concentration inequalities for the moment of the sum of functions of independent variables \cite{DBLP:conf/colt/BousquetKZ20}, which also has been employed for stability analysis in \cite{DBLP:conf/nips/LeiLK20, DBLP:conf/nips/LeiLY21}.

\begin{lemma}
\label{Lemma 2 of theorem 1}
\cite{DBLP:conf/colt/BousquetKZ20} For independent random variables with $z_i\in \mathcal{Z}, i\in [n]$, denote $S = \{z_1, ..., z_{n}\}$, $S \backslash \{z_i\}=\{z_1, ..., z_{i-1}, z_{i+1}, ..., z_{n}\}$ and functions $g_i: \mathcal{Z}^n \rightarrow \mathbb{R}, i\in [n]$. 
Suppose that for any $i\in [n]$:
\begin{itemize}
    \item $|\mathbb{E}_{S \backslash \{z_i\}}g_i(S)| \leq M$ almost surely (a.s.) with $M > 0$,
    \item $\mathbb{E}_{z_i}g_i(S) = 0$ a.s.,
    \item the difference of function $g_i$ can be bounded by $\beta$, i.e., $|g_i(z_1, ..., z_n) - g_i(z_1, ..., z_{j-1}, \bar{z}_{j}, z_{j+1}, ..., z_n)| \leq \beta$,
    where $j\in [n]$ with $i \neq j$, and $\bar{z}_j\in \mathcal{Z}$.
\end{itemize}
Then, 
\begin{equation*}
\begin{aligned}
    \Big|\Big|\sum\limits_{i=1}^{n} g_i(S)\Big|\Big|_p \leq 12\sqrt{2}pn\beta \lceil \mathrm{log}_2n \rceil + 4M\sqrt{pn}, \forall p\geq 2.
\end{aligned}
\end{equation*}
\end{lemma}

\begin{lemma}
\label{Lemma 3 of theorem 1}
\cite{DBLP:conf/colt/BousquetKZ20} If $\|Z\|_p \leq \sqrt{p}a + pb$, where $Z$ is a random variable, $p\geq 2$ and $a, b\in \mathbb{R}_+$. 
Then, for any $\delta \in (0, 1/e)$, we have 
\begin{equation*}
\begin{aligned}
    |Z| \leq e\Big(a\sqrt{\mathrm{log}(e/\delta)} + b\mathrm{log}(e/\delta)\Big)
\end{aligned}
\end{equation*}
with probability at least $1 - \delta$, where $e$ denotes the base of the natural logarithm.
\end{lemma}

The proof of Theorem 1 is obtained by integrating Lemmas 1, \ref{Lemma 2 of theorem 1}, \ref{Lemma 3 of theorem 1} with the fine-grained error decomposition. 

\begin{theorem1proof}
According to the definitions of $R_S(A(S))$ and $R(A(S))$, we know
\begin{equation*} 
\label{proof-th1-1}
\begin{aligned}
    |R(A(S)) - R_S(A(S))| 
    = \Bigg| \mathbb{E}_{z^+, \tilde{z}^+, z^-}\ell(A(S); z^+, \tilde{z}^+, z^-) - 
    \frac{1}{n_+(n_+-1)n_-} \sum\limits_{i, j\in [n_+], i\neq j, \atop k\in [n_-]}\ell(A(S); z_i^+, z_j^+, z_k^-) \Bigg|.
\end{aligned}
\end{equation*}
For convenience, we multiply this equation by $n_+ (n_+-1) n_-$ and then decompose it as follows
\begin{equation*}
\label{proof-th1-2}
\begin{aligned}
    & \Bigg| n_+(n_+-1)n_- \mathbb{E}_{z^+, \tilde{z}^+, z^-}\ell(A(S); z^+, \tilde{z}^+, z^-) - \sum\limits_{i, j\in [n_+], i\neq j, \atop k\in [n_-]}\ell(A(S); z_i^+, z_j^+, z_k^-) \Bigg| \nonumber\\
    = & \Bigg| \sum\limits_{i, j\in [n_+], i\neq j, \atop k\in [n_-]}\Big[\mathbb{E}_{z^+, \tilde{z}^+, z^-}\ell(A(S); z^+, \tilde{z}^+, z^-) - \ell(A(S); z_i^+, z_j^+, z_k^-)\Big] \Bigg| \nonumber\\
    = & \Bigg| \sum\limits_{i, j\in [n_+], i\neq j, \atop k\in [n_-]}\Big[\mathbb{E}_{z^+, \tilde{z}^+, z^-}\ell(A(S); z^+, \tilde{z}^+, z^-) - \mathbb{E}_{\bar{z}_i^+, \bar{z}_j^+, \bar{z}_k^-}\ell(A(S); z_i^+, z_j^+, z_k^-)\Big] \Bigg| \nonumber\\
    \leq & \Bigg| \sum\limits_{i, j\in [n_+], i\neq j, \atop k\in [n_-]} \mathbb{E}_{z^+, \tilde{z}^+, z^-}\Big[\ell(A(S); z^+, \tilde{z}^+, z^-) - \mathbb{E}_{\bar{z}_i^+, \bar{z}_j^+, \bar{z}_k^-}\ell(A(S_{i, j, k}); z^+, \tilde{z}^+, z^-) \Big] \Bigg| \\
    & + \Bigg| \sum\limits_{i, j\in [n_+], i\neq j, \atop k\in [n_-]} \mathbb{E}_{\bar{z}_i^+, \bar{z}_j^+,  \bar{z}_k^-} \Big[ \mathbb{E}_{z^+, \tilde{z}^+, z^-} \ell(A(S_{i, j, k}); z^+, \tilde{z}^+, z^-) - \ell(A(S_{i, j, k}); z_i^+, z_j^+, z_k^-) \Big] \Bigg| \\
    & + \Bigg| \sum\limits_{i, j\in [n_+], i\neq j, \atop k\in [n_-]} \mathbb{E}_{\bar{z}_i^+, \bar{z}_j^+,  \bar{z}_k^-} [\ell(A(S_{i, j, k}); z_i^+, z_j^+, z_k^-) - \ell(A(S); z_i^+, z_j^+, z_k^-)] \Bigg|,
\end{aligned}
\end{equation*}
where the definition of  $S_{i, j, k}$ is given in Definition 2 (also see the proof of Lemma 1). 

Now we try to bound the three parts of the above decomposition. For the first and third parts, according to Lemma 1, we deduce that 
\begin{equation*}\label{proof-th1-3}
\begin{aligned}
    & \Bigg| \sum\limits_{i, j\in [n_+], i\neq j, \atop k\in [n_-]} \mathbb{E}_{z^+, \tilde{z}^+, z^-}\Big[\ell(A(S); z^+, \tilde{z}^+, z^-) - \mathbb{E}_{\bar{z}_i^+, \bar{z}_j^+, \bar{z}_k^-}\ell(A(S_{i, j, k}); z^+, \tilde{z}^+, z^-) \Big] \Bigg| \nonumber\\
    \leq & \Bigg| n_+(n_+-1)n_- \mathop{\mathrm{sup}}\limits_{z^+, \tilde{z}^+\in \mathcal{Z}_+, \atop z^-\in \mathcal{Z}_-} \Big[\ell(A(S); z^+, \tilde{z}^+, z^-) - \mathbb{E}_{\bar{z}_i^+, \bar{z}_j^+, \bar{z}_k^-}\ell(A(S_{i, j, k}); z^+, \tilde{z}^+, z^-) \Big] \Bigg| \nonumber\\
    \leq & 3n_+(n_+-1)n_-\gamma
\end{aligned}
\end{equation*}
and
\begin{equation}\label{proof-th1-4}
\begin{aligned}
    & \Bigg| \sum\limits_{i, j\in [n_+], i\neq j, \atop k\in [n_-]} \mathbb{E}_{\bar{z}_i^+, \bar{z}_j^+,  \bar{z}_k^-} [\ell(A(S_{i, j, k}); z_i^+, z_j^+, z_k^-) - \ell(A(S); z_i^+, z_j^+, z_k^-)] \Bigg| \nonumber\\
    \leq & \Bigg| n_+(n_+-1)n_- \mathop{\mathrm{sup}}\limits_{\bar{z}_i^+, \bar{z}_j^+\in \mathcal{Z}_+, \atop \bar{z}_k^-\in \mathcal{Z}_-} \big(\ell(A(S_{i, j, k}); z_i^+, z_j^+, z_k^-) - \ell(A(S); z_i^+, z_j^+, z_k^-)\big) \Bigg| \nonumber\\
    \leq & 3n_+(n_+-1)n_-\gamma.
\end{aligned}
\end{equation}
Turn to the bound of the second part, let
\begin{equation*}
\begin{aligned}
     \Bigg| \sum\limits_{i, j\in [n_+], i\neq j, \atop k\in [n_-]} \mathbb{E}_{\bar{z}_i^+, \bar{z}_j^+,  \bar{z}_k^-} \Big[ \mathbb{E}_{z^+, \tilde{z}^+, z^-} \ell(A(S_{i, j, k}); z^+, \tilde{z}^+, z^-) - \ell(A(S_{i, j, k}); z_i^+, z_j^+, z_k^-) \Big] \Bigg|=\Bigg|\sum\limits_{i,j\in[n_+], i\neq j, \atop k\in [n_-]} g_{i, j, k}(S)\Bigg|,
\end{aligned}
\end{equation*}
where
\begin{equation*}
\begin{aligned}
    g_{i, j, k}(S) = & \mathbb{E}_{\bar{z}_i^+, \bar{z}_j^+,  \bar{z}_k^-} \Big[ \mathbb{E}_{z^+, \tilde{z}^+, z^-} \ell(A(S_{i, j, k}); z^+, \tilde{z}^+, z^-) - \ell(A(S_{i, j, k}); z_i^+, z_j^+, z_k^-) \Big]\\
    = & \mathbb{E}_{\bar{z}_i^+, \bar{z}_j^+,  \bar{z}_k^-} \Big[ \mathbb{E}_{z^+, \tilde{z}^+, z^-} \ell(A(S_{i, j, k}); z^+, \tilde{z}^+, z^-) - \mathbb{E}_{z^+, \tilde{z}^+} \ell(A(S_{i, j, k}); z^+, \tilde{z}^+, z_k^-)\\ 
    & + \mathbb{E}_{z^+, \tilde{z}^+} \ell(A(S_{i, j, k}); z^+, \tilde{z}^+, z_k^-) - \ell(A(S_{i, j, k}); z_i^+, z_j^+, z_k^-) \Big]\\
    = & \mathbb{E}_{\bar{z}_i^+, \bar{z}_j^+,  \bar{z}_k^-} \Big[ \mathbb{E}_{z^+, \tilde{z}^+, z^-} \ell(A(S_{i, j, k}); z^+, \tilde{z}^+, z^-) - \mathbb{E}_{z^+, \tilde{z}^+} \ell(A(S_{i, j, k}); z^+, \tilde{z}^+, z_k^-)\\ 
    & + \mathbb{E}_{z^+, \tilde{z}^+} \ell(A(S_{i, j, k}); z^+, \tilde{z}^+, z_k^-) - \mathbb{E}_{z^+} \ell(A(S_{i, j, k}); z^+, z_j^+, z_k^-) \\
    & + \mathbb{E}_{z^+} \ell(A(S_{i, j, k}); z^+, z_j^+, z_k^-) - \ell(A(S_{i, j, k}); z_i^+, z_j^+, z_k^-) \Big]\\
    := & g_k^{(i, j)}(S) + g_j^{(i, k)}(S) + g_i^{(j, k)}(S)
  \end{aligned}
\end{equation*}
with 
\begin{equation*}
\begin{aligned}
g_k^{(i, j)}(S):=  \mathbb{E}_{\bar{z}_i^+, \bar{z}_j^+,  \bar{z}_k^-} \Big[ \mathbb{E}_{z^+, \tilde{z}^+, z^-} \ell(A(S_{i, j, k}); z^+, \tilde{z}^+, z^-) - \mathbb{E}_{z^+, \tilde{z}^+} \ell(A(S_{i, j, k}); z^+, \tilde{z}^+, z_k^-)  \Big],\\ 
g_j^{(i, k)}(S):= \mathbb{E}_{\bar{z}_i^+, \bar{z}_j^+,  \bar{z}_k^-} \Big[ \mathbb{E}_{z^+, \tilde{z}^+} \ell(A(S_{i, j, k}); z^+, \tilde{z}^+, z_k^-) - \mathbb{E}_{z^+} \ell(A(S_{i, j, k}); z^+, z_j^+, z_k^-) \Big], \\
g_i^{(j, k)}(S):=\mathbb{E}_{\bar{z}_i^+, \bar{z}_j^+,  \bar{z}_k^-} \Big[ \mathbb{E}_{z^+} \ell(A(S_{i, j, k}); z^+, z_j^+, z_k^-) - \ell(A(S_{i, j, k}); z_i^+, z_j^+, z_k^-) \Big].
  \end{aligned}
\end{equation*}

For any fixed $i,j\in [n_+], i\neq j$, and consider $n_-$ random functions $g_1^{(i, j)}, ..., g_{n_-}^{(i, j)}$.
In terms of $|\mathbb{E}_{S}\ell(A(S); z^+, \tilde{z}^+, z^-)|\leq M, \forall z^+, \tilde{z}^+ \in \mathcal{Z}_+, z^-\in\mathcal{Z}_-$, we know
\begin{equation*}
\begin{aligned}
    |\mathbb{E}_{S \backslash \{z_k^-\}}g_k^{(i, j)}(S)| &= \Bigg| \mathbb{E}_{S \backslash \{z_k^-\}} \Bigg[ \mathbb{E}_{\bar{z}_i^+, \bar{z}_j^+, \bar{z}_k^-} \Big[ \mathbb{E}_{z^+, \tilde{z}^+, z^-} \ell(A(S_{i, j, k}); z^+, \tilde{z}^+, z^-) - \mathbb{E}_{z^+, \tilde{z}^+} \ell(A(S_{i, j, k}); z^+, \tilde{z}^+, z_k^-)  \Big] \Bigg] \Bigg|\\
    &= \Bigg| \mathbb{E}_{z_i^+, z_j^+} \Bigg[ \mathbb{E}_{S_{i, j, k}} \Big[ \mathbb{E}_{z^+, \tilde{z}^+, z^-} \ell(A(S_{i, j, k}); z^+, \tilde{z}^+, z^-) - \mathbb{E}_{z^+, \tilde{z}^+} \ell(A(S_{i, j, k}); z^+, \tilde{z}^+, z_k^-)  \Big] \Bigg] \Bigg|\\
    &\leq 2M.
\end{aligned}
\end{equation*}
Since $z_k^-$ is independent of $S_{i, j, k}$, we have
\begin{equation*}
\begin{aligned}
    |\mathbb{E}_{z_k^-}g_k^{(i, j)}(S)| &= \Bigg| \mathbb{E}_{z_k^-} \Bigg[ \mathbb{E}_{\bar{z}_i^+, \bar{z}_j^+, \bar{z}_k^-} \Big[ \mathbb{E}_{z^+, \tilde{z}^+, z^-} \ell(A(S_{i, j, k}); z^+, \tilde{z}^+, z^-) - \mathbb{E}_{z^+, \tilde{z}^+} \ell(A(S_{i, j, k}); z^+, \tilde{z}^+, z_k^-)  \Big] \Bigg] \Bigg|\\
    &= \Bigg| \mathbb{E}_{\bar{z}_i^+, \bar{z}_j^+, \bar{z}_k^-} \Bigg[ \mathbb{E}_{z_k^-} \Big[ \mathbb{E}_{z^+, \tilde{z}^+, z^-} \ell(A(S_{i, j, k}); z^+, \tilde{z}^+, z^-) - \mathbb{E}_{z^+, \tilde{z}^+} \ell(A(S_{i, j, k}); z^+, \tilde{z}^+, z_k^-)  \Big] \Bigg] \Bigg|\\
    &= \Bigg| \mathbb{E}_{\bar{z}_i^+, \bar{z}_j^+, \bar{z}_k^-} \Big[ \mathbb{E}_{z^+, \tilde{z}^+, z^-} \ell(A(S_{i, j, k}); z^+, \tilde{z}^+, z^-) - \mathbb{E}_{z^+, \tilde{z}^+, z_k^-} \ell(A(S_{i, j, k}); z^+, \tilde{z}^+, z_k^-) \Big] \Bigg|\\
    &= 0.
\end{aligned}
\end{equation*}
For any $m\in [n_-]$ and $\bar{z}_m^- \in \mathcal{Z}_-$, we can easily check that
\begin{equation*}
\begin{aligned}
    &|g_k^{(i, j)}(S) - g_k^{(i, j)}(S^{(m)})| \\
    = & \Bigg| \mathbb{E}_{\bar{z}_i^+, \bar{z}_j^+, \bar{z}_k^-} \Big[ \mathbb{E}_{z^+, \tilde{z}^+, z^-} \ell(A(S_{i, j, k}); z^+, \tilde{z}^+, z^-) - \mathbb{E}_{z^+, \tilde{z}^+} \ell(A(S_{i, j, k}); z^+, \tilde{z}^+, z_k^-)  \Big]\\
    & - \mathbb{E}_{\bar{z}_i^+, \bar{z}_j^+, \bar{z}_k^-} \Big[ \mathbb{E}_{z^+, \tilde{z}^+, z^-} \ell(A(S_{i, j, k}^{(m)}); z^+, \tilde{z}^+, z^-) - \mathbb{E}_{z^+, \tilde{z}^+} \ell(A(S_{i, j, k}^{(m)}); z^+, \tilde{z}^+, z_k^-)  \Big] \Bigg|\\
    \leq & \Bigg| \mathbb{E}_{\bar{z}_i^+, \bar{z}_j^+, \bar{z}_k^-} \Big[ \mathbb{E}_{z^+, \tilde{z}^+, z^-} [\ell(A(S_{i, j, k}); z^+, \tilde{z}^+, z^-) - \ell(A(S_{i, j, k}^{(m)}); z^+, \tilde{z}^+, z^-)] \Big] \Bigg|\\
    & + \Bigg| \mathbb{E}_{\bar{z}_i^+, \bar{z}_j^+, \bar{z}_k^-} \Big[ \mathbb{E}_{z^+, \tilde{z}^+} [\ell(A(S_{i, j, k}); z^+, \tilde{z}^+, z_k^-) - \ell(A(S_{i, j, k}^{(m)}); z^+, \tilde{z}^+, z_k^-)]  \Big] \Bigg|\\
    \leq & 2\gamma,
\end{aligned}
\end{equation*}
where $S^{(m)}$ and $S_{i, j, k}^{(m)}$ are respect to  $S$ and $S_{i, j, k}$ after  replacing $z_m^-$ with $\bar{z}_m^-$,  respectively.
Therefore, we prove that, for $g_1^{(i, j)}, ..., g_{n_-}^{(i, j)}$, all the assumptions of Lemma \ref{Lemma 2 of theorem 1} hold with $n$ replaced by $n_-$ and $\beta = 2\gamma$. 
Based on Lemma \ref{Lemma 2 of theorem 1}, for any $i, j \in [n_+], i\neq j$, we get
\begin{equation*}
\begin{aligned}
    \Big|\Big|\sum\limits_{k\in [n_-]}g_k^{(i, j)}(S)\Big|\Big|_p \leq 24\sqrt{2}pn_-\gamma \lceil \mathrm{log}_2n_- \rceil + 8M\sqrt{pn_-}.
\end{aligned}
\end{equation*}
Similarly, we also have
\begin{equation*}
\begin{aligned}
    \Big|\Big|\sum\limits_{j\in [n_+], i\neq j}g_j^{(i, k)}(S)\Big|\Big|_p \leq 24\sqrt{2}p(n_+-1)\gamma \lceil \mathrm{log}_2(n_+-1) \rceil + 8M\sqrt{p(n_+-1)}, \quad \forall i\in [n_+], k\in [n_-]
\end{aligned}
\end{equation*}
and
\begin{equation*}
\begin{aligned}
    \Big|\Big|\sum\limits_{i\in [n_+], i\neq j}g_i^{(j, k)}(S)\Big|\Big|_p \leq 24\sqrt{2}p(n_+-1)\gamma \lceil \mathrm{log}_2(n_+-1) \rceil + 8M\sqrt{p(n_+-1)}, \quad \forall j\in [n_+], k\in [n_-].
\end{aligned}
\end{equation*}
Then, by direct computation, we derive that
\begin{equation*}
\begin{aligned}
    &\Bigg|\Bigg|\sum\limits_{i,j\in[n_+], i\neq j, \atop k\in [n_-]} g_{i, j, k}(S)\Bigg|\Bigg|_p
    = \Bigg|\Bigg|\sum\limits_{i,j\in[n_+], i\neq j, \atop k\in [n_-]} \Big(g_k^{(i, j)}(S) + g_j^{(i, k)}(S) + g_i^{(j, k)}(S)\Big)\Bigg|\Bigg|_p\\
    \leq & \Bigg|\Bigg|\sum\limits_{i,j\in[n_+], i\neq j, \atop k\in [n_-]} g_{k}^{(i, j)}(S)\Bigg|\Bigg|_p + \Bigg|\Bigg|\sum\limits_{i,j\in[n_+], i\neq j, \atop k\in [n_-]} g_{j}^{(i, k)}(S)\Bigg|\Bigg|_p + \Bigg|\Bigg|\sum\limits_{i,j\in[n_+], i\neq j, \atop k\in [n_-]} g_{i}^{(j, k)}(S)\Bigg|\Bigg|_p\\
    \leq & \sum\limits_{i, j\in[n_+], \atop i\neq j} \Bigg|\Bigg|\sum\limits_{k\in [n_-]} g_{k}^{(i, j)}(S)\Bigg|\Bigg|_p + \sum\limits_{i\in [n_+], \atop  k\in [n_-]} \Bigg|\Bigg|\sum\limits_{j\in[n_+], i\neq j} g_{j}^{(i, k)}(S)\Bigg|\Bigg|_p + \sum\limits_{j\in [n_+], \atop  k\in [n_-]} \Bigg|\Bigg|\sum\limits_{i\in[n_+], i\neq j} g_{i}^{(j, k)}(S)\Bigg|\Bigg|_p\\
    \leq & 24\sqrt{2}pn_+(n_+-1)n_-\gamma \lceil \mathrm{log}_2n_-\rceil + 8Mn_+(n_+-1)\sqrt{pn_-}\\
    & + \big(24\sqrt{2}pn_+(n_+-1)n_-\gamma \lceil \mathrm{log}_2(n_+-1)\rceil + 8Mn_+n_-\sqrt{p(n_+-1)}\big) \times 2\\
    = & 24\sqrt{2}pn_+(n_+-1)n_-\gamma [\lceil \mathrm{log}_2n_-\rceil + 2\lceil \mathrm{log}_2(n_+-1)\rceil] + 8Mn_+ \big((n_+-1)\sqrt{pn_-} + 2n_-\sqrt{p(n_+-1)}\big)\\
    \leq & 24\sqrt{2}pn_+(n_+-1)n_-\gamma \Big(\lceil \mathrm{log}_2\big(n_-(n_+-1)^2\big)\rceil + 2 \Big) + 8Mn_+\sqrt{p}\Big((n_+-1)\sqrt{n_-} + 2n_-\sqrt{n_+-1} \Big),
\end{aligned}
\end{equation*}
where the first and second inequalities are built with the subadditivity of $\|\cdot\|_p$. 
Hence, according to Lemma \ref{Lemma 3 of theorem 1}, the third error term 
\begin{equation*}
\begin{aligned}
    \Bigg|\sum\limits_{i,j\in[n_+], i\neq j, \atop k\in [n_-]} g_{i, j, k}(S)\Bigg| 
    \leq & e\Bigg(8Mn_+\Big((n_+-1)\sqrt{n_-} + 2n_-\sqrt{n_+-1} \Big) \sqrt{\mathrm{log}(e/\delta)}\\
    & + 24\sqrt{2}n_+(n_+-1)n_-\gamma \Big(\lceil \mathrm{log}_2\big(n_-(n_+-1)^2\big)\rceil + 2\Big) \mathrm{log}(e/\delta) \Bigg).
\end{aligned}
\end{equation*}
The desired result follows by combining the estimations of the three error terms. 
\end{theorem1proof}

\subsection{B.2~~~ Proof of Theorem 2}

Following the analysis in \cite{DBLP:conf/icml/HardtRS16}, we can verify the following property for the gradient update. 

\begin{lemma}
\label{Lemma 2 of theorem 2}
   Assume that the loss function $w \rightarrow \ell(w; z^+, \tilde{z}^+, z^-)$ is convex and $\alpha$-smooth, where $z^+, \tilde{z}^+\in \mathcal{Z}_+, z^-\in \mathcal{Z}_-$. Then, for any $\eta \leq 2/\alpha$, the optimization process of SGD is $1$-expansive, that is
    \begin{equation*}
    \begin{aligned}
        \|w - \eta \nabla\ell(w; z^+, \tilde{z}^+, z^-) - w^{\prime} + \eta \nabla\ell(w^{\prime}; z^+, \tilde{z}^+, z^-)\|
        \leq \|w - w^{\prime}\|.
    \end{aligned}
    \end{equation*}
\end{lemma}

\begin{lemma2proof}
Assume $S$ and $S^{\prime}$ are two datasets that differ only by the last example among the former $n_+$ samples.
If $i_t, j_t\in [n_+-1], i_t\neq j_t$ and $k_t\in [n_-]$, according to (3) and Lemma \ref{Lemma 2 of theorem 2}, we know that
\begin{equation*}
\begin{aligned}
    \|w_{t+1} - w_{t+1}^{\prime}\| &= \|w_t - \eta_t\nabla\ell(w_t; z_{i_t}^+, z_{j_t}^+, z_{k_t}^-) - w_t^{\prime} + \eta_t\nabla\ell(w_t^{\prime}; \bar{z}_{i_t}^+, \bar{z}_{j_t}^+, \bar{z}_{k_t}^-)\|\\
    &= \|w_t - \eta_t\nabla\ell(w_t; z_{i_t}^+, z_{j_t}^+, z_{k_t}^-) - w_t^{\prime} + \eta_t\nabla\ell(w_t^{\prime}; z_{i_t}^+, z_{j_t}^+, z_{k_t}^-)\|\\
    &\leq \|w_t - w_t^{\prime}\|.
\end{aligned}
\end{equation*}
If $i_t = n_+$ or $j_t = n_+$, $i_t \neq j_t$, $k_t\in [n_-]$, we get that
\begin{equation*}
\begin{aligned}
    \|w_{t+1} - w_{t+1}^{\prime}\| &= \|w_t - \eta_t\nabla\ell(w_t; z_{i_t}^+, z_{j_t}^+, z_{k_t}^-) - w_t^{\prime} + \eta_t\nabla\ell(w_t^{\prime}; \bar{z}_{i_t}^+, \bar{z}_{j_t}^+, \bar{z}_{k_t}^-)\|\\
    &\leq \|w_t - w_t^{\prime}\| + \|\eta_t\nabla\ell(w_t; \bar{z}_{i_t}^+, \bar{z}_{j_t}^+, \bar{z}_{k_t}^-) - \eta_t\nabla\ell(w_t^{\prime}; z_{i_t}^+, z_{j_t}^+, z_{k_t}^-)\|\\
    &\leq \|w_t - w_t^{\prime}\| + 2\eta_t L.
\end{aligned}
\end{equation*}
Assume $S$ and $S^{\prime}$ are two datasets that differ only by the last example among the later $n_-$ samples.
If $i_t, j_t\in [n_+]$, $i_t \neq j_t$ and $k_t \in [n_--1]$, we have
\begin{equation*}
\begin{aligned}
    \|w_{t+1} - w_{t+1}^{\prime}\| &= \|w_t - \eta_t\nabla\ell(w_t; z_{i_t}^+, z_{j_t}^+, z_{k_t}^-) - w_t^{\prime} + \eta_t\nabla\ell(w_t^{\prime}; \bar{z}_{i_t}^+, \bar{z}_{j_t}^+, \bar{z}_{k_t}^-)\|\\
    &= \|w_t - \eta_t\nabla\ell(w_t; z_{i_t}^+, z_{j_t}^+, z_{k_t}^-) - w_t^{\prime} + \eta_t\nabla\ell(w_t^{\prime}; z_{i_t}^+, z_{j_t}^+, z_{k_t}^-)\|\\
    &\leq \|w_t - w_t^{\prime}\|.
\end{aligned}
\end{equation*}
Similarly, for $i_t, j_t\in [n_+]$, $i_t \neq j_t$ and $k_t = n_-$, there holds
\begin{equation*}
\begin{aligned}
    \|w_{t+1} - w_{t+1}^{\prime}\| &= \|w_t - \eta_t\nabla\ell(w_t; z_{i_t}^+, z_{j_t}^+, z_{k_t}^-) - w_t^{\prime} + \eta_t\nabla\ell(w_t^{\prime}; z_{i_t}^+, z_{j_t}^+, \bar{z}_{k_t}^-)\|\\
    &\leq \|w_t - w_t^{\prime}\| + \|\eta_t\nabla\ell(w_t; z_{i_t}^+, z_{j_t}^+, \bar{z}_{k_t}^-) - \eta_t\nabla\ell(w_t^{\prime}; z_{i_t}^+, z_{j_t}^+, z_{k_t}^-)\|\\
    &\leq \|w_t - w_t^{\prime}\| + 2\eta_t L.
\end{aligned}
\end{equation*}
As a combination of the above four cases, we derive that
\begin{equation*}
\begin{aligned}
    & \|w_{t + 1} - w_{t + 1}^{\prime}\|\\
    \leq & \|w_t - w_t^{\prime}\| + 2\eta_t L\mathbb{I}\Big[(i_t = n_+ ~\mathrm{or}~ j_t = n_+, i_t \neq j_t, k_t\in [n_-], z_{n_+}^+\neq \bar{z}_{n_+}^+) ~\mathrm{or}~ (i_t, j_t\in [n_+], i_t \neq j_t, k_t = n_-, z_{n_-}^-\neq \bar{z}_{n_-}^-) \Big]\\
    \leq & 2L\sum\limits_{l = 1}^{t}\eta_l \mathbb{I}\Big[(i_l = n_+ ~\mathrm{or}~ j_l = n_+, i_l \neq j_l, k_l\in [n_-], z_{n_+}^+\neq \bar{z}_{n_+}^+) ~\mathrm{or}~ (i_l, j_l\in [n_+], i_l \neq j_l, k_l = n_-, z_{n_-}^-\neq \bar{z}_{n_-}^-) \Big].
\end{aligned}
\end{equation*}
Finally, according to the  $L$-Lipschitz of the loss function $\ell(w; z^+, \tilde{z}^+, z^-)$, we deduce that
\begin{equation*}
\begin{aligned}
    & |\ell(w_{t + 1}; z^+, \tilde{z}^+, z^-) - \ell(w_{t + 1}^{\prime}; z^+, \tilde{z}^+, z^-)|\\
    \leq & \mathop{\mathrm{sup}}\limits_{z^+, \tilde{z}^+\in \mathcal{Z}_+, \atop z^-\in \mathcal{Z}_-} |\ell(w_{t + 1}; z^+, \tilde{z}^+, z^-) - \ell(w_{t + 1}^{\prime}; z^+, \tilde{z}^+, z^-)| \leq L\|w_{t + 1} - w_{t + 1}^{\prime}\|\\
    \leq & 2L^2\sum\limits_{l = 1}^t \eta_l \mathbb{I}\Big[(i_l = n_+ ~\mathrm{or}~ j_l = n_+, i_l \neq j_l, k_l\in [n_-], z_{n_+}^+\neq \bar{z}_{n_+}^+) ~\mathrm{or}~ (i_l, j_l\in [n_+], i_l \neq j_l, k_l = n_-, z_{n_-}^-\neq \bar{z}_{n_-}^-) \Big].
\end{aligned}
\end{equation*}

This completes the desired result. 
\end{lemma2proof}

Lemma 2 implies that the difference between two model sequences will not increase with the number of iteration under proper conditions. 
Usually, a sufficiently small iteration step size also plays an important role on further limiting its change and guaranteeing  the stability.

The Chernoff's bound described as below is used in our proof.

\begin{lemma}
\label{Lemma 3 of theorem 2}
    \cite{DBLP:books/daglib/BoucheronLM13}
    Let $X = \sum\limits_{t=1}^{T}X_t$ and $\mu = \mathbb{E}X$, where $X_1, ..., X_T$ be independent Bernoulli random variables. 
    Then, for any $\delta \in (0, 1)$, 
    \begin{equation*}
    \begin{aligned}
        X \leq (1 + \sqrt{3\mathrm{log}(1/\delta)/\mu})\mu
    \end{aligned}
    \end{equation*}
   with probability at least $1 - \delta$.
\end{lemma}

\begin{theorem2proof}
Lemma 2 assures that SGD with $t$-iterations is $\gamma$-uniformly stable with
\begin{equation*}
\begin{aligned}
    \gamma \leq 2L^2\sum\limits_{l = 1}^t \eta_l \mathbb{I}\Big[(i_l = n_+ ~\mathrm{or}~ j_l = n_+, i_l \neq j_l, k_l\in [n_-], z_{n_+}^+\neq \bar{z}_{n_+}^+) ~\mathrm{or}~ (i_l, j_l\in [n_+], i_l \neq j_l, k_l = n_-, z_{n_-}^-\neq \bar{z}_{n_-}^-) \Big].
\end{aligned}
\end{equation*}
Let $A(S) = w_T$. From Theorem 1, we know the following inequality holds with probability $1 - \delta$
\begin{equation*}
\begin{aligned}
    |R_S(w_T) - R(w_T)| \leq 6\gamma + 
    e\Bigg(8M(\frac{1}{\sqrt{n_-}} + \frac{2}{\sqrt{n_+ - 1}})\sqrt{\mathrm{log}(e/\delta)} + 24\sqrt{2}\gamma \Big(\lceil \mathrm{log}_2\big(n_-(n_+ - 1)^2\big) \rceil + 2\Big) \mathrm{log}(e/\delta)\Bigg).
\end{aligned}
\end{equation*}
Thus,
\begin{equation}
\label{equation 1 of theorem 2}
\begin{aligned}
    |R_S(w_T) - R(w_T)| = O\Bigg(
    & \Big(\lceil \mathrm{log}\big(n_-(n_+ - 1)^2\big) \rceil + 2\Big) \mathrm{log}(1/\delta) \sum\limits_{t = 1}^T \eta \mathbb{I}\Big[(i_t = n_+ ~\mathrm{or}~ j_t = n_+,  i_t \neq j_t, k_t\in [n_-], z_{n_+}^+\neq \bar{z}_{n_+}^+)\\
    & ~\mathrm{or}~ (i_t, j_t\in [n_+], i_t \neq j_t, k_t = n_-, z_{n_-}^-\neq \bar{z}_{n_-}^-) \Big] + \Big(\frac{1}{\sqrt{n_-}} + \frac{2}{\sqrt{n_+-1}} \Big) \sqrt{\mathrm{log}(1/\delta)} \Bigg),
\end{aligned}
\end{equation}
where $\eta = \eta_t \equiv c/\sqrt{T}, c\leq 2/\alpha$.

Let 
$$X_t = \mathbb{I}\Big[(i_t = n_+ ~\mathrm{or}~ j_t = n_+, i_t \neq j_t, k_t\in [n_-], z_{n_+}^+\neq \bar{z}_{n_+}^+) ~\mathrm{or}~ (i_t, j_t\in [n_+], i_t \neq j_t, k_t = n_-, z_{n_-}^-\neq \bar{z}_{n_-}^-) \Big]$$ and $\mu = \sum\limits_{t = 1}^T\mathbb{E}X_t$.
It is easy to verify that
\begin{equation*}
\begin{aligned}
    \mathbb{E}X_t &= \mathrm{Prob}\Big\{\mathbb{I}\Big[(i_t = n_+ ~\mathrm{or}~ j_t = n_+, i_t \neq j_t, k_t\in [n_-], z_{n_+}^+\neq \bar{z}_{n_+}^+) ~\mathrm{or}~ (i_t, j_t\in [n_+], i_t \neq j_t, k_t = n_-, z_{n_-}^-\neq \bar{z}_{n_-}^-) \Big] \Big\}\\
    & \leq \mathrm{Prob}\Big\{i_t = n_+ ~\mathrm{or}~ j_t = n_+, i_t \neq j_t, k_t\in [n_-], z_{n_+}^+\neq \bar{z}_{n_+}^+ \Big\} + \mathrm{Prob}\Big\{i_t, j_t\in [n_+], i_t \neq j_t, k_t = n_-, z_{n_-}^-\neq \bar{z}_{n_-}^- \Big\}\\
    & = \frac{1}{n_+} + \frac{1}{2n_-}.
\end{aligned}
\end{equation*}
Meanwhile,
\begin{equation*}
\begin{aligned}
    \mathbb{E}X_t \geq \mathrm{Prob}\Big\{i_t = n_+ ~\mathrm{or}~ j_t = n_+, i_t \neq j_t, k_t\in [n_-], z_{n_+}^+\neq \bar{z}_{n_+}^+ \Big\}
\end{aligned}
\end{equation*}
and
\begin{equation*}
\begin{aligned}
    \mathbb{E}X_t \geq \mathrm{Prob}\Big\{i_t, j_t\in [n_+], i_t \neq j_t, k_t = n_-, z_{n_-}^-\neq \bar{z}_{n_-}^- \Big\},
\end{aligned}
\end{equation*}
that is,
\begin{equation*}
\begin{aligned}
    \mathbb{E}X_t \geq \mathrm{max}\Big\{\frac{1}{n_+}, \frac{1}{2n_-}\Big\}.
\end{aligned}
\end{equation*}
It follows that
\begin{equation*}
\begin{aligned}
    \mathrm{max}\Big\{\frac{T}{n_+}, \frac{T}{2n_-}\Big\} \leq \mu = \sum\limits_{t = 1}^T\mathbb{E}X_t \leq \frac{T}{n_+} + \frac{T}{2n_-}.
\end{aligned}
\end{equation*}
Applying Lemma \ref{Lemma 3 of theorem 2} with 
$$X_t = \mathbb{I}\Big[(i_t = n_+ ~\mathrm{or}~ j_t = n_+, i_t \neq j_t, k_t\in [n_-], z_{n_+}^+\neq \bar{z}_{n_+}^+) ~\mathrm{or}~ (i_t, j_t\in [n_+], i_t \neq j_t, k_t = n_-, z_{n_-}^-\neq \bar{z}_{n_-}^-) \Big],$$ we have
\begin{equation}
\label{equation 2 of theorem 2}
\begin{aligned}
    X = \sum\limits_{t = 1}^T X_t \leq \Bigg(1 + \sqrt{\frac{3}{\mu}\mathrm{log}(1/\delta)} \Bigg)\mu \leq \Bigg(1 + \sqrt{\frac{3\mathrm{log}(1/\delta)}{\mathrm{max}\{\frac{T}{n_+}, \frac{T}{2n_-}\}}} \Bigg) \Big(\frac{T}{n_+} + \frac{T}{2n_-} \Big)
\end{aligned}
\end{equation}
with probability $1 - \delta$. 
Combining \eqref{equation 1 of theorem 2} and  \eqref{equation 2 of theorem 2}, with probability $1 - \delta$,  we have
\begin{equation*}
\begin{aligned}
    &|R_S(w_T) - R(w_T)| \\
    = & O\Bigg(\Big(\lceil \mathrm{log}\big(n_-(n_+ - 1)^2\big) \rceil + 2\Big) \mathrm{log}(1/\delta) \sum\limits_{t = 1}^T \eta \mathbb{I}\Big[(i_t = n_+ ~\mathrm{or}~ j_t = n_+, i_t \neq j_t, k_t\in [n_-], z_{n_+}^+\neq \bar{z}_{n_+}^+)\\
    & ~\mathrm{or}~ (i_t, j_t\in [n_+], i_t \neq j_t, k_t = n_-, z_{n_-}^-\neq \bar{z}_{n_-}^-) \Big] + \Big(\frac{1}{\sqrt{n_-}} + \frac{2}{\sqrt{n_+-1}} \Big) \sqrt{\mathrm{log}(1/\delta)} \Bigg)\\
    = & O\Bigg(\Big(\lceil \mathrm{log}\big(n_-(n_+ - 1)^2\big) \rceil + 2\Big) \mathrm{log}(1/\delta) \eta \sum\limits_{t = 1}^T X_t + \Big(\frac{1}{\sqrt{n_-}} + \frac{2}{\sqrt{n_+-1}} \Big) \sqrt{\mathrm{log}(1/\delta)} \Bigg)\\
    \leq & O\Bigg(\Big(\lceil \mathrm{log}\big(n_-(n_+ - 1)^2\big) \rceil + 2\Big) \mathrm{log}(1/\delta) \eta \Big(1 + \sqrt{\frac{3\mathrm{log}(1/\delta)}{\mathrm{max}\{\frac{T}{n_+}, \frac{T}{2n_-}\}}} \Big) \Big(\frac{T}{n_+} + \frac{T}{2n_-} \Big)\\
    &+ \Big(\frac{1}{\sqrt{n_-}} + \frac{2}{\sqrt{n_+-1}} \Big) \sqrt{\mathrm{log}(1/\delta)} \Bigg)\\
    = & O\Bigg(\Big(\lceil \mathrm{log}\big(n_-(n_+ - 1)^2\big) \rceil + 2\Big) \mathrm{log}(1/\delta) \Big(1 + \sqrt{\frac{\mathrm{log}(1/\delta)}{\mathrm{max}\{\frac{T}{n_+}, \frac{T}{n_-}\}}} \Big) \Big(\frac{\sqrt{T}}{n_+} + \frac{\sqrt{T}}{n_-} \Big)\\
    &+ \Big(\frac{1}{\sqrt{n_-}} + \frac{1}{\sqrt{n_+-1}} \Big) \sqrt{\mathrm{log}(1/\delta)} \Bigg).
\end{aligned}
\end{equation*}

This proves the desired statement. 
\end{theorem2proof}

\begin{lemma3proof}
Because the triplet learning involves two different sample spaces, some related works \cite{DBLP:conf/nips/LeiLK20, ArXiv:Pitcan17} can not be used here directly. Fortunately, by detailed decomposition, we have 
\begin{equation*}
\begin{aligned}
    &\Bigg|\frac{1}{n_+(n_+-1)n_-} \sum\limits_{i,j\in [n_+], i \neq j, \atop k\in [n_-]} \ell(w^*; z_i^+, z_j^+, z_k^-) - \mathbb{E}_{z^+, \tilde{z}^+, z^-}\ell(w^*; z^+, \tilde{z}^+, z^-) \Bigg|\\
    \leq &\Bigg|\frac{2}{n_+(n_+-1)n_-} \sum\limits_{i,j\in [n_+], i \neq j, \atop k\in [n_-]} \ell(w^*; z_i^+, z_j^+, z_k^-) - 2\mathbb{E}_{z^+, \tilde{z}^+, z^-}\ell(w^*; z^+, \tilde{z}^+, z^-) \Bigg|\\
    \leq &\Bigg|\frac{1}{n_+(n_+-1)n_-} \sum\limits_{k\in [n_-]} \sum\limits_{i,j\in [n_+], i \neq j} \ell(w^*; z_i^+, z_j^+, z_k^-) - \mathbb{E}_{z^-} \mathbb{E}_{z^+, \tilde{z}^+}\ell(w^*; z^+, \tilde{z}^+, z^-)\Bigg|\\
    &+ \Bigg|\frac{1}{n_+(n_+-1)n_-} \sum\limits_{i,j\in [n_+], i \neq j} \sum\limits_{k\in [n_-]} \ell(w^*; z_i^+, z_j^+, z_k^-) - \mathbb{E}_{z^+, \tilde{z}^+} \mathbb{E}_{z^-} \ell(w^*; z^+, \tilde{z}^+, z^-)\Bigg|\\
    \leq &\Bigg| \mathop{\mathrm{sup}}_{k\in [n_-]} \Bigg( \frac{1}{n_+(n_+-1)} 
    \sum\limits_{i,j\in [n_+], i \neq j} \ell(w^*; z_i^+, z_j^+) - \mathbb{E}_{z^+, \tilde{z}^+}\ell(w^*; z^+, \tilde{z}^+) \Bigg) \Bigg|\\
    &+ \Bigg| \mathop{\mathrm{sup}}_{i,j\in [n_+], i \neq j} \Bigg( \frac{1}{n_-} \sum\limits_{k\in [n_-]} \ell(w^*; z_k^-) - \mathbb{E}_{z^-} \ell(w^*; z^-) \Bigg) \Bigg|\\
    \leq &\frac{2b\mathrm{log}(1/\delta)}{3\lfloor n_+/2 \rfloor} +\sqrt{\frac{2\tau\mathrm{log}(1/\delta)}{\lfloor n_+/2 \rfloor}} + \frac{2b\mathrm{log}(1/\delta)}{3\lfloor n_- \rfloor} + \sqrt{\frac{2\tau\mathrm{log}(1/\delta)}{\lfloor n_- \rfloor}},
\end{aligned}
\end{equation*}
where the second inequality follows from triangular inequality and and the last inequality is derived by combining Lemma B.3 \cite{DBLP:conf/nips/LeiLK20} and Theorem 2.4  \cite{ArXiv:Pitcan17}.

This completes the proof.
\end{lemma3proof}

\subsection{B.3~~~ Proof of Theorem 3 and Supplement to Remark 5}
\begin{lemma4proof}
Let $$S_m = \{z_1^+, ..., z_{m-1}^+, \bar{z}_m^+, z_{m+1}^+, ..., z_{n+}^+, z_1^-, ..., z_{n_-}^-\} ~\mathrm{or}~ \{z_1^+, ..., z_{n+}^+, z_1^-, ..., z_{m-1}^-, \bar{z}_m^-, z_{m+1}^-, ..., z_{n_-}^-\}.$$
Due to the generality of $m$, we can firstly assume $S_m = S_{n_+} = \{ z_1^+, ..., z_{n_+-1}^+, \bar{z}_{n_+}^+, z_1^-, ..., z_{n_-}^- \}$.
Since $A(S_{n_+})$ is a minimizer of $F_{S_{n_+}}(w)$, we know
\begin{equation*}
\begin{aligned}
    F_S(A(S_{n_+})) - F_S(A(S)) & = F_S(A(S_{n_+})) - F_{S_{n_+}}(A(S_{n_+})) + F_{S_{n_+}}(A(S_{n_+})) - F_{S_{n_+}}(A(S)) + F_{S_{n_+}}(A(S)) - F_S(A(S))\\
    & \leq F_S(A(S_{n_+})) - F_{S_{n_+}}(A(S_{n_+})) + F_{S_{n_+}}(A(S)) - F_S(A(S)).
\end{aligned}
\end{equation*}
By the definitions of $F_S$ and $F_{S_{n_+}}$, we further get 
\begin{equation*}
\begin{aligned}
    & n_+(n_+-1)n_-\Big( F_S(A(S_{n_+})) - F_{S_{n_+}}(A(S_{n_+})) \Big)\\
    = & \sum\limits_{i,j \in [n_+], i\neq j, \atop k\in [n_-]} f(A(S_{n_+}); z_i^+, z_j^+, z_k^-) - \Bigg( \sum\limits_{i,j \in [n_+-1], i\neq j, \atop k\in [n_-]} f(A(S_{n_+}); z_i^+, z_j^+, z_k^-) + \sum\limits_{i \in [n_+-1], \atop k\in [n_-]} f(A(S_{n_+}); z_i^+, \bar{z}_{n_+}^+, z_k^-)\\
    & + \sum\limits_{j \in [n_+-1], \atop k\in [n_-]} f(A(S_{n_+}); \bar{z}_{n_+}^+, z_j^+, z_k^-) \Bigg)\\
    = & \sum\limits_{i\in [n_+-1], \atop k\in [n_-]}\Big( f(A(S_{n_+}); z_i^+, z_{n_+}^+, z_k^-) + f(A(S_{n_+}); z_{n_+}^+, z_i^+, z_k^-) - f(A(S_{n_+}); z_i^+, \bar{z}_{n_+}^+, z_k^-) - f(A(S_{n_+}); \bar{z}_{n_+}^+, z_i^+, z_k^-) \Big), 
\end{aligned}
\end{equation*}
where
$$f(w; z^+, \tilde{z}^+, z^-)=\ell(w; z^+, \tilde{z}^+, z^-)+r(w).$$
Similarly, 
\begin{equation*}
\begin{aligned}
    & n_+(n_+-1)n_-\Big( F_{S_{n_+}}(A(S)) - F_{S}(A(S)) \Big)\\
    = & \sum\limits_{i\in [n_+-1], \atop k\in [n_-]}\Big( f(A(S); z_i^+, \bar{z}_{n_+}^+, z_k^-) + f(A(S); \bar{z}_{n_+}^+, z_i^+, z_k^-) - f(A(S); z_i^+, z_{n_+}^+, z_k^-) - f(A(S); z_{n_+}^+, z_i^+, z_k^-) \Big).
\end{aligned}
\end{equation*}
Based on the above quantitative relations, we get
\begin{equation*}
\begin{aligned}
    F_S(A(S_{n_+})) - F_S(A(S)) 
    \leq \frac{1}{n_+(n_+-1)n_-} \sum\limits_{i\in [n_+-1], \atop k\in [n_-]}
    & \Big(\big(\ell(A(S_{n_+}); z_i^+, z_{n_+}^+, z_k^-)-\ell(A(S); z_i^+, z_{n_+}^+, z_k^-)\big)\\
    & + \big(\ell(A(S_{n_+}); z_{n_+}^+, z_i^+, z_k^-)-\ell(A(S); z_{n_+}^+, z_i^+, z_k^-)\big)\\
    & + \big(\ell(A(S); z_i^+, \bar{z}_{n_+}^+, z_k^-)-\ell(A(S_{n_+}); z_i^+, \bar{z}_{n_+}^+, z_k^-)\big)\\
    & + \big(\ell(A(S); \bar{z}_{n_+}^+, z_i^+, z_k^-)-\ell(A(S_{n_+}); \bar{z}_{n_+}^+, z_i^+, z_k^-)\big)\Big).
\end{aligned}
\end{equation*}
With the same analysis as above, we also can obtain that 
\begin{equation*}
\begin{aligned}
    F_S(A(S_{n_-})) - F_S(A(S)) 
    \leq \frac{1}{n_+(n_+-1)n_-} \sum\limits_{i,j\in [n_+], i\neq j}
    & \Big(\big(\ell(A(S_{n_-}); z_i^+, z_j^+, z_{n_-}^-)-\ell(A(S); z_i^+, z_j^+, z_{n_-}^-)\big)\\
    & + \big(\ell(A(S); z_i^+, z_j^+, \bar{z}_{n_-}^-)-\ell(A(S_{n_-}); z_i^+, z_j^+, \bar{z}_{n_-}^-)\big)\Big)
\end{aligned}
\end{equation*}
when $S_m = S_{n_-} = \{ z_1^+, ..., z_{n_+}^+, z_1^-, ..., z_{n_--1}^-, \bar{z}_{n_-}^- \}$.

Moreover, for  $A(S) = \arg \min \limits_{w\in \mathcal{W}} F_S(w)$,
\begin{equation*}
\begin{aligned}
    F_S(A(S_m)) - F_S(A(S)) \leq \frac{1}{n_+(n_+-1)n_-} \sum\limits_{i\in [n_+], i\neq m, \atop k\in [n_-]} 
    & \Big(\big( \ell(A(S_m); z_i^+, z_m^+, z_k^-) - \ell(A(S); z_i^+, z_m^+, z_k^-) \big) \\
    & +
    \big( \ell(A(S_m); z_m^+, z_i^+, z_k^-) - \ell(A(S); z_m^+, z_i^+, z_k^-) \big) \\
    & + \big( \ell(A(S); z_i^+, \bar{z}_m^+, z_k^-) - \ell(A(S_m); z_i^+, \bar{z}_m^+, z_k^-) \big) \\
    & + \big( \ell(A(S); \bar{z}_m^+, z_i^+, z_k^-) - \ell(A(S_m); \bar{z}_m^+, z_i^+, z_k^-) \big)\Big)
\end{aligned}
\end{equation*}
as  
$S_m = \{ z_1^+, ..., z_{m - 1}^+, \bar{z}_m^+, z_{m + 1}^+, ..., z_{n_+}^+, z_1^-, ..., z_{n_-}^- \}$ and  
\begin{equation*}
\begin{aligned}
    F_S(A(S_m)) - F_S(A(S)) \leq \frac{1}{n_+(n_+-1)n_-}\sum\limits_{i,j\in [n_+], i\neq j} 
    & \Big(\big( \ell(A(S_m); z_i^+, z_j^+, z_m^-) - \ell(A(S); z_i^+, z_j^+, z_m^-) \big) \\
    & +
    \big( \ell(A(S); z_i^+, z_j^+, \bar{z}_m^-) - \ell(A(S_m); z_i^+, z_j^+, \bar{z}_m^-) \big)\Big)
\end{aligned}
\end{equation*}
as $S_m = \{ z_1^+, ..., z_{n_+}^+, z_1^-, ..., z_{m - 1}^-, \bar{z}_m^-, z_{m + 1}^-, ..., z_{n_-}^- \}$.
  
Following the similar proof steps of  Lemma B.2 \cite{DBLP:conf/nips/LeiLK20}, we can get the desired result with the help of 
 the $\sigma$-strong convexity of $F_S(w)$ and the $L$-Lipschitz continuity of $\ell(w)$.
\end{lemma4proof}

\begin{lemma5proof}
Lemma 4 tells us that  $A$ is $\gamma$-uniformly stable with $\gamma = \mathrm{min} \Big\{\frac{8}{n_+}, \frac{4}{n_-}\Big\} \frac{L^2}{\sigma}$. 
Similar with Lemma 2 in \citet{DBLP:conf/nips/LeiLK20}, there holds $$\mathbb{E}_S\|A(S) - w^*\|^2 \leq 12\gamma/\sigma = \mathrm{min}\Big\{\frac{96}{n_+},\frac{48}{n_-}\Big\}\frac{L^2}{\sigma^2}.$$
Based on the Cauchy-Schwartz inequality, we further have 
\begin{equation}
\label{equation 1 of theorem 3}
\begin{aligned}
    \mathbb{E}_S\|A(S) - w^*\| \leq \sqrt{\mathbb{E}_S\|A(S) - w^*\|^2} \leq \mathrm{min}\Big\{\frac{4\sqrt{6}}{\sqrt{n_+}},\frac{4\sqrt{3}}{\sqrt{n_-}}\Big\}\frac{L}{\sigma}.
\end{aligned}
\end{equation}
It is easy to verify that
\begin{equation*}
\begin{aligned}
    & \mathop{\mathrm{sup}}\limits_{z^+, \tilde{z}^+\in \mathcal{Z}_+, \atop z^-\in \mathcal{Z}_-} |\ell(A(S); z^+, \tilde{z}^+, z^-) - \ell(A(\bar{S}); z^+, \tilde{z}^+, z^-)|\\
    = & \mathop{\mathrm{sup}}\limits_{z^+, \tilde{z}^+\in \mathcal{Z}_+, \atop z^-\in \mathcal{Z}_-} |\ell(A(S); z^+, \tilde{z}^+, z^-) - \ell(w^*; z^+, \tilde{z}^+, z^-) - \ell(A(\bar{S}); z^+, \tilde{z}^+, z^-) + \ell(w^*; z^+, \tilde{z}^+, z^-)|\\
    \leq & \mathrm{min}\Big\{\frac{8}{n_+},\frac{4}{n_-}\Big\}\frac{L^2}{\sigma}.
\end{aligned}
\end{equation*}

Let $\tilde{\ell}(w; z^+, \tilde{z}^+, z^-) = \ell(w; z^+, \tilde{z}^+, z^-) - \ell(w^*; z^+, \tilde{z}^+, z^-), \forall z^+, \tilde{z}^+\in \mathcal{Z}_+, z^-\in \mathcal{Z}_-, w\in \mathcal{W}$.
Then, the algorithm $A$ measured by the loss function $\tilde{\ell}$ is also $\mathrm{min} \Big\{ \frac{8}{n_+},\frac{4}{n_-}\Big\}\frac{L^2}{\sigma}$-uniformly stable.
Besides, 
\begin{equation*}
\begin{aligned}
    |\mathbb{E}_S\tilde{\ell}(A(S); z^+, \tilde{z}^+, z^-)| & = |\mathbb{E}_S[\ell(A(S); z^+, \tilde{z}^+, z^-) - \ell(w^*; z^+, \tilde{z}^+, z^-)]|\\
    & \leq \mathbb{E}_S|\ell(A(S); z^+, \tilde{z}^+, z^-) - \ell(w^*; z^+, \tilde{z}^+, z^-)|\\
    & \leq L\mathbb{E}_S\|A(S) - w^*\|\\
    & \leq \mathrm{min}\Big\{ \frac{4\sqrt{6}}{\sqrt{n_+}}, \frac{4\sqrt{3}}{\sqrt{n_-}}\Big\}\frac{L^2}{\sigma},
\end{aligned}
\end{equation*}
where the last two inequalities are built from the $L$-Lipschitz continuity and \eqref{equation 1 of theorem 3}, respectively.
\end{lemma5proof}

\begin{theorem3proof}
Recall that $A$ measured by the loss function $\tilde{\ell}$ is $\mathrm{min} \Big\{\frac{8}{n_+}, \frac{4}{n_-}\Big\}\frac{L^2}{\sigma}$-uniformly stable and $|\mathbb{E}_S\tilde{\ell}(A(S); z^+, \tilde{z}^+, z^-)| \leq M = \mathrm{min}\Big\{ \frac{4\sqrt{6}}{\sqrt{n_+}}, \frac{4\sqrt{3}}{\sqrt{n_-}}\Big\}\frac{L^2}{\sigma}$.
According to Theorem 1, we have with probability $1 - \delta$
\begin{equation*}
\begin{aligned}
    & |R_S(A(S)) - R(A(S))|\\
    \leq & \Bigg| \frac{1}{n_+(n_+-1)n_-} \sum\limits_{i,j\in[n_+], i\neq j, \atop k\in [n_-]} \ell(w^*; z_i^+, z_j^+, z_k^-) - \mathbb{E}_{z^+, \tilde{z}^+, z^-}\ell(w^*; z^+, \tilde{z}^+, z^-) \Bigg|\\
    & + \Bigg| \frac{1}{n_+(n_+-1)n_-} \sum\limits_{i,j\in[n_+], i\neq j, \atop k\in [n_-]} \tilde{\ell}(A(S); z_i^+, z_j^+, z_k^-) - \mathbb{E}_{z^+, \tilde{z}^+, z^-}\tilde{\ell}(A(S); z^+, \tilde{z}^+, z^-) \Bigg|\\
    \leq & \Bigg| \frac{1}{n_+(n_+-1)n_-} \sum\limits_{i,j\in[n_+], i\neq j, \atop k\in [n_-]} \ell(w^*; z_i^+, z_j^+, z_k^-) - \mathbb{E}_{z^+, \tilde{z}^+, z^-}\ell(w^*; z^+, \tilde{z}^+, z^-) \Bigg| + \mathrm{min}\Big\{ \frac{48}{n_+}, \frac{24}{n_-}\Big\}\frac{L^2}{\sigma} \\ 
    & +\frac{eL^2}{\sigma}\Bigg( \mathrm{min}\Big\{ \frac{32\sqrt{6}}{\sqrt{n_+}}, \frac{32\sqrt{3}}{\sqrt{n_-}}\Big\} \Big( \frac{1}{\sqrt{n_-}}+\frac{2}{\sqrt{n_+-1}} \Big)\sqrt{\mathrm{log}(e/\delta)} + \mathrm{min}\Big\{ \frac{192\sqrt{2}}{n_+}, \frac{96\sqrt{2}}{n_-}\Big\}  \Big( \lceil \mathrm{log}_2\big(n_-(n_+-1)^2\big) \rceil + 2 \Big) \mathrm{log}(e/\delta) \Bigg).
\end{aligned}
\end{equation*}
Lemma 3 assures that, with probability at least $1 - \delta$, 
\begin{equation*}
\begin{aligned}
    & \Bigg|\frac{1}{n_+(n_+-1)n_-} \sum\limits_{i,j\in [n_+], i \neq j, \atop k\in [n_-]} \ell(w^*; Z_i^+, Z_j^+, Z_k^-) - \mathbb{E}_{z^+, \tilde{z}^+, z^-}\ell(w^*; z^+, \tilde{z}^+, z^-) \Bigg|\\
    \leq & \frac{2b\mathrm{log}(1/\delta)}{3\lfloor n_+/2 \rfloor} + \sqrt{\frac{2\tau\mathrm{log}(1/\delta)}{\lfloor n_+/2 \rfloor}} + \frac{2b\mathrm{log}(1/\delta)}{3\lfloor n_- \rfloor} + \sqrt{\frac{2\tau\mathrm{log}(1/\delta)}{\lfloor n_- \rfloor}}.
\end{aligned}
\end{equation*}
Then,
\begin{equation*}
\begin{aligned}
    & |R_S(A(S)) - R(A(S))|\\
    \leq & \frac{2b\mathrm{log}(1/\delta)}{3\lfloor n_+/2 \rfloor} + \sqrt{\frac{2\tau\mathrm{log}(1/\delta)}{\lfloor n_+/2 \rfloor}} + \frac{2b\mathrm{log}(1/\delta)}{3\lfloor n_- \rfloor} + \sqrt{\frac{2\tau\mathrm{log}(1/\delta)}{\lfloor n_- \rfloor}} + \mathrm{min}\Big\{ \frac{48}{n_+}, \frac{24}{n_-}\Big\}\frac{L^2}{\sigma} \\
    & +\frac{eL^2}{\sigma}\Bigg(\mathrm{min}\Big\{ \frac{32\sqrt{6}}{\sqrt{n_+}}, \frac{32\sqrt{3}}{\sqrt{n_-}}\Big\} \Big( \frac{1}{\sqrt{n_-}}+\frac{2}{\sqrt{n_+-1}} \Big)\sqrt{\mathrm{log}(e/\delta)} + \mathrm{min}\Big\{ \frac{192\sqrt{2}}{n_+}, \frac{96\sqrt{2}}{n_-}\Big\} \Big( \lceil \mathrm{log}_2\big(n_-(n_+-1)^2\big) \rceil + 2 \Big) \mathrm{log}(e/\delta) \Bigg)\\
    = & O\Bigg( \sqrt{\frac{\mathrm{log}(1/\delta)}{n_+}} + \sqrt{\frac{\mathrm{log}(1/\delta)}{n_-}} + \sigma^{-1} \Big(\mathrm{min}\Big\{\frac{\sqrt{2}}{\sqrt{n_+}}, \frac{1}{\sqrt{n_-}}\Big\} \Big( \frac{1}{\sqrt{n_-}} + \frac{1}{\sqrt{n_+}}\Big) \sqrt{\mathrm{log}(1/\delta)}\\
    & + \mathrm{min}\Big\{ \frac{2}{n_+}, \frac{1}{n_-} \Big\} \mathrm{log}\big(n_-n_+^2\big)\mathrm{log}(1/\delta) \Big) \Bigg).
\end{aligned}
\end{equation*}
\end{theorem3proof}

\begin{remark5proof}
When letting $r(w^*) = \lambda \|w^*\|^2 = O(\sigma \|w^*\|^2)$, we can verify the strong convexity of $F_S$ as follows. Due to the convexity of $\ell$, we know that $R_S(w) \geq R_S(w^\prime) + \langle \nabla R_S(w^\prime), w - w^\prime\rangle$.
Besides,
\begin{equation*}
\begin{aligned}
    \lambda \|w\|^2
    & = \lambda \|w\|^2 + 2\lambda \|w^\prime\|^2 - 2\lambda \|w^\prime\|^2 + 2\lambda \langle w, w^\prime \rangle - 2\lambda \langle w, w^\prime \rangle\\
    & = \lambda \|w^\prime\|^2 + \langle 2\lambda \|w^\prime\|, w - w^\prime \rangle + \lambda (\|w\|^2 + \|w^\prime\|^2 - 2\langle w, w^\prime \rangle)\\
    & = \lambda \|w^\prime\|^2 + \langle 2\lambda \|w^\prime\|, w - w^\prime \rangle + \lambda \|w - w^\prime\|^2.
\end{aligned}
\end{equation*}
Combining the above inequality and identity, we derive that
\begin{equation*}
\begin{aligned}
    R_S(w) + \lambda \|w\|^2 \geq R_S(w^\prime) + \lambda \|w^\prime\|^2 + \langle \nabla R_S(w^\prime) + \nabla \lambda \|w^\prime\|^2, w - w^\prime\rangle + \lambda \|w - w^\prime\|^2,
\end{aligned}
\end{equation*}
that is,
\begin{equation}
\label{regularization}
\begin{aligned}
    F_S(w) \geq F_S(w^\prime) + \langle \nabla F_S(w^\prime), w - w^\prime\rangle + \lambda \|w - w^\prime\|^2.
\end{aligned}
\end{equation}
Thus, $F_S$ is $2\lambda$-strongly convex, i.e. $\sigma = 2\lambda$.

\end{remark5proof}

\subsection{B.4~~~Proof of Theorem 4 and Supplement to Remark 6}
\begin{lemma6proof}
As illustrated in \citet{DBLP:conf/nips/SrebroST10}, the $\alpha$-smooth and non-negative function $\ell$ satisfies 
\begin{equation}
\label{equation 1 of lemma 11}
\begin{aligned}
    \|\ell^\prime (w)\|^2 \leq 2\alpha \ell(w).
\end{aligned}
\end{equation}
In addition, from the convexity and $\alpha$-smoothness of $\ell$, we can also derive that
\begin{equation}
\label{equation 2 of lemma 11}
\begin{aligned}
    \ell(w) \leq \ell(w^\prime) + \langle \nabla \ell(w^\prime), w - w^\prime \rangle + \frac{\alpha\|w - w^\prime\|^2}{2}, \forall w, w^\prime \in \mathcal{W},
\end{aligned}
\end{equation}
where $\nabla \ell$ denotes subgradient of $\ell$.
Based on \eqref{equation 1 of lemma 11}, \eqref{equation 2 of lemma 11} and the Cauchy-Schwartz inequality, we deduce that 
\begin{equation*}
\begin{aligned}
    & \ell(A(S_{i,j,k}); z_i^+, z_j^+, z_k^-) - \ell(A(S); z_i^+, z_j^+, z_k^-)\\
    \leq & \langle \nabla\ell(A(S); z_i^+, z_j^+, z_k^-), A(S_{i,j,k}) - A(S) \rangle + \frac{\alpha}{2} \|A(S_{i,j,k}) - A(S)\|^2\\
    \leq & \|\nabla\ell(A(S); z_i^+, z_j^+, z_k^-)\| \|A(S_{i,j,k}) - A(S)\| + \frac{\alpha}{2} \|A(S_{i,j,k}) - A(S)\|^2\\
    \leq & \frac{1}{2\epsilon} \|\nabla\ell(A(S); z_i^+, z_j^+, z_k^-)\|^2 + \frac{\epsilon}{2} \|A(S_{i,j,k}) - A(S)\|^2 + \frac{\alpha}{2} \|A(S_{i,j,k}) - A(S)\|^2\\
    \leq & \frac{\alpha}{\epsilon} \ell(A(S); z_i^+, z_j^+, z_k^-) + \frac{\epsilon + \alpha}{2} \|A(S_{i,j,k}) - A(S)\|^2.
\end{aligned}
\end{equation*}

Moreover, according to the definition of triplet on-average stability in Definition 2 and the symmetry of  $A(S_{i,j,k})$ w.r.t. $z_i^+, z_j^+, z_k^-$, we have
\begin{equation*}
\begin{aligned}
    & \mathbb{E}_S[R(A(S)) - R_S(A(S))]\\
    = & \frac{1}{n_+(n_+-1)n_-} \sum\limits_{i,j\in [n_+], i\neq j, \atop k\in [n_-]} \mathbb{E}_{S, \bar{S}}[R(A(S_{i,j,k})) - R_S(A(S))]\\
    = & \frac{1}{n_+(n_+-1)n_-} \sum\limits_{i,j\in [n_+], i\neq j, \atop k\in [n_-]} \mathbb{E}_{S, \bar{S}}[\ell(A(S_{i,j,k}); z_i^+, z_j^+, z_k^-) - \ell(A(S); z_i^+, z_j^+, z_k^-)]\\
    \leq & \frac{1}{n_+(n_+-1)n_-} \sum\limits_{i,j\in [n_+], i\neq j, \atop k\in [n_-]} \mathbb{E}_{S, \bar{S}}\Big[\frac{\alpha}{\epsilon} \ell(A(S); z_i^+, z_j^+, z_k^-) + \frac{\epsilon + \alpha}{2} \|A(S_{i,j,k}) - A(S)\|^2\Big]\\
    = & \frac{\alpha}{\epsilon} \mathbb{E}_SR_S(A(S)) + \frac{\epsilon + \alpha}{2n_+(n_+-1)n_-} \sum\limits_{i,j\in [n_+], i\neq j, \atop k\in [n_-]} \mathbb{E}_{S, \bar{S}} \|A(S_{i,j,k}) - A(S)\|^2\\
    = & \frac{\alpha}{\epsilon} \mathbb{E}_SR_S(A(S)) + \frac{\epsilon + \alpha}{2n_+(n_+-1)n_-} \sum\limits_{i,j\in [n_+], i\neq j, \atop k\in [n_-]} \mathbb{E}_{S, \bar{S}} \|A(S_{i,j,k}) - A(S_{i,j}) + A(S_{i,j}) -A(S_i) + A(S_i) - A(S)\|^2\\
    \leq & \frac{\alpha}{\epsilon} \mathbb{E}_SR_S(A(S)) + \frac{\epsilon + \alpha}{2n_+(n_+-1)n_-} \sum\limits_{i,j\in [n_+], i\neq j, \atop k\in [n_-]} \Big(3\mathbb{E}_{S, \bar{S}} \|A(S_{i,j,k}) - A(S_{i,j})\|^2 + 3\mathbb{E}_{S, \bar{S}} \|A(S_{i,j}) -A(S_i)\|^2\\
    & + 3\mathbb{E}_{S, \bar{S}} \|A(S_i) - A(S)\|^2\Big)\\
    = & \frac{\alpha}{\epsilon} \mathbb{E}_SR_S(A(S)) + \frac{3(\epsilon + \alpha)}{2n_+(n_+-1)n_-} \sum\limits_{i\in [n_+], \atop k\in [n_-]} \Big(\mathbb{E}_{S, \bar{S}} \|A(S_k) - A(S)\|^2 + 2\mathbb{E}_{S, \bar{S}}\|A(S_i) -A(S)\|^2\Big).
\end{aligned}
\end{equation*}

The desired result is proved.

\end{lemma6proof}

\begin{theorem4proof}
Similar with the proof of Lemma 4, when $S_m = \{ z_1^+, ..., z_{m-1}^+, \bar{z}_m^+, z_{m+1}^+, ..., z_{n_+}^+, z_1^-, ..., z_{n_-}^- \}$, we have
\begin{equation*}
\begin{aligned}
    & n_+(n_+-1)n_-\Big( F_S(A(S_m)) - F_S(A(S)) \Big)\\
    \leq & \sum\limits_{i\in [n_+], i\neq m, \atop k\in [n_-]} \Big(\big( \ell(A(S_m); z_i^+, z_m^+, z_k^-) - \ell(A(S); z_i^+, z_m^+, z_k^-) \big) + \big( \ell(A(S_m); z_m^+, z_i^+, z_k^-) - \ell(A(S); z_m^+, z_i^+, z_k^-) \big)\\
    & + \big( \ell(A(S); z_i^+, \bar{z}_m^+, z_k^-) - \ell(A(S_m); z_i^+, \bar{z}_m^+, z_k^-) \big) + \big( \ell(A(S); \bar{z}_m^+, z_i^+, z_k^-) - \ell(A(S_m); \bar{z}_m^+, z_i^+, z_k^-) \big)\Big)\\
    \leq & \sum\limits_{i\in [n_+], i\neq m, \atop k\in [n_-]} \Big( \big\langle \nabla\ell(A(S); z_i^+, z_m^+, z_k^-) + \nabla\ell(A(S); z_m^+, z_i^+, z_k^-) - \nabla\ell(A(S_m); z_i^+, \bar{z}_m^+, z_k^-) - \nabla\ell(A(S_m); \bar{z}_m^+, z_i^+, z_k^-),\\
    & A(S_m) - A(S)\big\rangle + 2\alpha \|A(S_m) - A(S)\|^2 \Big)\\
    \leq & \sum\limits_{i\in [n_+], i\neq m, \atop k\in [n_-]} \Big( \|\nabla\ell(A(S); z_i^+, z_m^+, z_k^-)\| + \|\nabla\ell(A(S); z_m^+, z_i^+, z_k^-)\| + \|\nabla\ell(A(S_m); z_i^+, \bar{z}_m^+, z_k^-)\| + \|\nabla\ell(A(S_m); \bar{z}_m^+, z_i^+, z_k^-)\| \Big)\\
    & \|A(S_m) - A(S)\| + 2\alpha (n_+-1)n_- \|A(S_m) - A(S)\|^2,
\end{aligned}
\end{equation*}
where the second inequality is due to \eqref{equation 2 of lemma 11}.

Moreover, considering the definition of $A(S)$ and the $\sigma$-strong convexity of $F$, we can derive
\begin{equation*}
\begin{aligned}
    & \frac{\sigma n_+(n_+-1)n_-}{2} \|A(S_m) - A(S)\|^2\\
    \leq & n_+(n_+-1)n_- \Big( F_S(A(S_m)) - F_S(A(S)) \Big)\\
    \leq & \sum\limits_{i\in [n_+], i\neq m, \atop k\in [n_-]} \Big( \|\nabla\ell(A(S); z_i^+, z_m^+, z_k^-)\| + \|\nabla\ell(A(S); z_m^+, z_i^+, z_k^-)\| + \|\nabla\ell(A(S_m); z_i^+, \bar{z}_m^+, z_k^-)\| + \|\nabla\ell(A(S_m); \bar{z}_m^+, z_i^+, z_k^-)\| \Big)\\
    & \|A(S_m) - A(S)\| + 2\alpha (n_+-1)n_- \|A(S_m) - A(S)\|^2\\
    \leq & \sqrt{2\alpha} \sum\limits_{i\in [n_+], i\neq m, \atop k\in [n_-]} \Big( \sqrt{\ell(A(S); z_i^+, z_m^+, z_k^-)} + \sqrt{\ell(A(S); z_m^+, z_i^+, z_k^-)} + \sqrt{\ell(A(S_m); z_i^+, \bar{z}_m^+, z_k^-)} + \sqrt{\ell(A(S_m); \bar{z}_m^+, z_i^+, z_k^-)} \Big)\\
    & \|A(S_m) - A(S)\| + 2\alpha (n_+-1)n_- \|A(S_m) - A(S)\|^2.
\end{aligned}
\end{equation*}
Then, as $8\alpha \leq \sigma \mathrm{min}\{n_+, n_-\} \leq \sigma n_+$, it follows that 
\begin{equation*}
\begin{aligned}
    & \frac{\sigma n_+(n_+-1)n_-}{2} \|A(S_m) - A(S)\|\\
    \leq & \sqrt{2\alpha} \sum\limits_{i\in [n_+], i\neq m, \atop k\in [n_-]} \Big( \sqrt{\ell(A(S); z_i^+, z_m^+, z_k^-)} + \sqrt{\ell(A(S); z_m^+, z_i^+, z_k^-)} + \sqrt{\ell(A(S_m); z_i^+, \bar{z}_m^+, z_k^-)} + \sqrt{\ell(A(S_m); \bar{z}_m^+, z_i^+, z_k^-)} \Big)\\
    & + 2\alpha (n_+-1)n_- \|A(S_m) - A(S)\|\\
    \leq & \sqrt{2\alpha} \sum\limits_{i\in [n_+], i\neq m, \atop k\in [n_-]} \Big( \sqrt{\ell(A(S); z_i^+, z_m^+, z_k^-)} + \sqrt{\ell(A(S); z_m^+, z_i^+, z_k^-)} + \sqrt{\ell(A(S_m); z_i^+, \bar{z}_m^+, z_k^-)} + \sqrt{\ell(A(S_m); \bar{z}_m^+, z_i^+, z_k^-)} \Big)\\
    & + \frac{\sigma n_+(n_+-1)n_-}{4} \|A(S_m) - A(S)\|.
\end{aligned}
\end{equation*}
That is to say
\begin{equation*}
\begin{aligned}
    & \frac{\sigma n_+(n_+-1)n_-}{4} \|A(S_m) - A(S)\|\\
    \leq & \sqrt{2\alpha} \sum\limits_{i\in [n_+], i\neq m, \atop k\in [n_-]} \Big( \sqrt{\ell(A(S); z_i^+, z_m^+, z_k^-)} + \sqrt{\ell(A(S); z_m^+, z_i^+, z_k^-)} + \sqrt{\ell(A(S_m); z_i^+, \bar{z}_m^+, z_k^-)} + \sqrt{\ell(A(S_m); \bar{z}_m^+, z_i^+, z_k^-)} \Big).
\end{aligned}
\end{equation*}
After squaring the both sides of the above inequality, we derive
\begin{equation*}
\begin{aligned}
    & \frac{\sigma^2 n_+^2(n_+-1)^2n_-^2}{16} \|A(S_m) - A(S)\|^2\\
    \leq & 2\alpha \Bigg( \sum\limits_{i\in [n_+], i\neq m, \atop k\in [n_-]} \Big( \sqrt{\ell(A(S); z_i^+, z_m^+, z_k^-)} + \sqrt{\ell(A(S); z_m^+, z_i^+, z_k^-)} + \sqrt{\ell(A(S_m); z_i^+, \bar{z}_m^+, z_k^-)} + \sqrt{\ell(A(S_m); \bar{z}_m^+, z_i^+, z_k^-)} \Big) \Bigg)^2\\
    \leq & 2\alpha (n_+-1)n_- \sum\limits_{i\in [n_+], i\neq m, \atop k\in [n_-]} \Big( \sqrt{\ell(A(S); z_i^+, z_m^+, z_k^-)} + \sqrt{\ell(A(S); z_m^+, z_i^+, z_k^-)} + \sqrt{\ell(A(S_m); z_i^+, \bar{z}_m^+, z_k^-)}\\
    & + \sqrt{\ell(A(S_m); \bar{z}_m^+, z_i^+, z_k^-)} \Big)^2\\
    \leq & 8\alpha (n_+-1)n_- \sum\limits_{i\in [n_+], i\neq m, \atop k\in [n_-]} \Big( \ell(A(S); z_i^+, z_m^+, z_k^-) + \ell(A(S); z_m^+, z_i^+, z_k^-) + \ell(A(S_m); z_i^+, \bar{z}_m^+, z_k^-) + \ell(A(S_m); \bar{z}_m^+, z_i^+, z_k^-) \Big),
\end{aligned}
\end{equation*}
where the second and third inequalities hold since $\Big(\sum\limits_{i=1}^n a_i\Big)^2 \leq n \sum\limits_{i=1}^n a_i^2$.

For all $m\in [n_+]$, we sum them together to get
\begin{equation*}
\begin{aligned}
    & \sigma^2 n_+^2(n_+-1)n_- \sum\limits_{m=1}^{n_+} \|A(S_m) - A(S)\|^2\\
    \leq & 128\alpha \sum\limits_{i, m\in [n_+], i\neq m, \atop k\in [n_-]} \Big( \ell(A(S); z_i^+, z_m^+, z_k^-) + \ell(A(S); z_m^+, z_i^+, z_k^-) + \ell(A(S_m); z_i^+, \bar{z}_m^+, z_k^-) + \ell(A(S_m); \bar{z}_m^+, z_i^+, z_k^-) \Big).
\end{aligned}
\end{equation*}
By taking expectations on both sides of the above inequality, we obtain
\begin{equation}
\label{equation 1 of theorem 4}
\begin{aligned}
    & \sigma^2 n_+^2(n_+-1)n_- \sum\limits_{m=1}^{n_+} \mathbb{E}_{S, \bar{S}} \|A(S_m) - A(S)\|^2\\
    \leq & 128\alpha \sum\limits_{i, m\in [n_+], i\neq m, \atop k\in [n_-]} \mathbb{E}_{S, \bar{S}} [ \ell(A(S); z_i^+, z_m^+, z_k^-) + \ell(A(S); z_m^+, z_i^+, z_k^-) + \ell(A(S_m); z_i^+, \bar{z}_m^+, z_k^-) + \ell(A(S_m); \bar{z}_m^+, z_i^+, z_k^-) ]\\
    \leq & 512\alpha n_+(n_+-1)n_-\mathbb{E}_{S} R_S(A(S)).
\end{aligned}
\end{equation}

Similarly, when $S_m = \{ z_1^+, ..., z_{n_+}^+, z_1^-, ..., z_{m-1}^-, \bar{z}_m^-, z_{m+1}^-, ..., z_{n_-}^- \}$, we get
\begin{equation}
\label{equation 2 of theorem 4}
\begin{aligned}
    \sigma^2 n_- \sum\limits_{m=1}^{n_-} \mathbb{E}_{S, \bar{S}}\|A(S_m) - A(S)\|^2 \leq \frac{512}{9} \alpha \mathbb{E}_SR_S(A(S)).
\end{aligned}
\end{equation}
Based on the definitions of $A(S)$ and $w^*$,  there holds
\begin{equation*}
\begin{aligned}
    \mathbb{E}_S[F(A(S)) - F_S(w^*)] & = \mathbb{E}_S[F(A(S)) - F_S(A(S))] + \mathbb{E}_S[F_S(A(S)) - F_S(w^*)]\\
    & \leq \mathbb{E}_S[F(A(S)) - F_S(A(S))] = \mathbb{E}_S[R(A(S)) - R_S(A(S))].
\end{aligned}
\end{equation*}  
Hence, the desired result 
\begin{equation*}
\begin{aligned}
    & \mathbb{E}_S[R(A(S)) - R_S(A(S))]\\
    \leq & \frac{\alpha}{\epsilon} \mathbb{E}_SR_S(A(S)) + \frac{3(\epsilon + \alpha)}{2n_+(n_+-1)n_-} \bigg( \frac{1024\alpha n_-}{\sigma^2 n_+} \mathbb{E}_SR_S(A(S)) + \frac{512\alpha n_+}{9\sigma^2 n_-} \mathbb{E}_SR_S(A(S)) \bigg)\\
    = & \bigg( \frac{\alpha}{\epsilon} + \frac{1536\alpha(\epsilon + \alpha)}{n_+^2(n_+-1)\sigma^2} + \frac{256\alpha (\epsilon + \alpha)}{3(n_+-1)n_-^2\sigma^2} \bigg) \mathbb{E}_SR_S(A(S)), \forall \epsilon > 0
\end{aligned}
\end{equation*}
follows from  Lemma 6, \eqref{equation 1 of theorem 4}, and \eqref{equation 2 of theorem 4}.
\end{theorem4proof}

\begin{remark6proof}
Considering (6) and letting $r(w^*) = O(\sigma \|w^*\|^2)$, we know
\begin{equation*}
\begin{aligned}
    \mathbb{E}_S[R(A(S)) - R(w^*)]
    & \leq \mathbb{E}_S[R(A(S)) - R_S(A(S)) + R_S(w^*) - R(w^*) +r(w^*)]\\
    & \leq \mathbb{E}_S[R(A(S)) - R_S(A(S)) + r(w^*)]\\
    & \leq \mathbb{E}_S[R(A(S)) - R_S(A(S))] + O(\sigma \|w^*\|^2).
\end{aligned}
\end{equation*}
If $n_+\asymp n_-\asymp n$ and $\epsilon = \sqrt{\frac{3 n_+^2 (n_+-1) n_-^2 \sigma^2}{4608 n_-^2 + 256 n_+^2}}$, then
\begin{equation*}
\begin{aligned}
    \mathbb{E}_S[R(A(S)) - R(w^*)] \leq \mathbb{E}_S[R(A(S)) - R_S(A(S))] + O(\sigma \|w^*\|^2) = O\bigg(\frac{1}{n^{\frac{3}{2}}\sigma} \mathbb{E}_SR_S(A(S)) + \sigma \|w^*\|^2\bigg).
\end{aligned}
\end{equation*}
By applying the non-negativity of $r$, the definitions of $A(S)$ and $w^*$, we get
\begin{equation*}
\begin{aligned}
    \mathbb{E}_SR_S(A(S)) \leq \mathbb{E}_S[R_S(A(S)) + r(A(S))] \leq \mathbb{E}_S[R_S(w^*) + r(w^*)] = O(R(w^*) + \sigma \|w^*\|^2).
\end{aligned}
\end{equation*}
Combining the above two inequalities together, we derive
\begin{equation*}
\begin{aligned}
    \mathbb{E}_S[R(A(S)) - R(w^*)] = O\bigg(\frac{1}{n^{\frac{3}{2}}\sigma} R(w^*) + \big(n^{-\frac{3}{2}} + \sigma \big)\|w^*\|^2 \bigg).
\end{aligned}
\end{equation*}
To minimize $\mathbb{E}_S[R(A(S)) - R(w^*)]$, $\sigma$ should be $n^{-\frac{3}{4}}\|w^*\|^{-1}\sqrt{R(w^*)}$, however, $\sigma \geq 8\alpha/n$.
When $R(w^*) = O(n^{-\frac{1}{2}}\|w^*\|^2)$, the order of $n^{-\frac{3}{4}}\|w^*\|^{-1}\sqrt{R(w^*)}$ is same as $8\alpha/n$. Thus, we derive 
\begin{equation*}
\begin{aligned}
    \mathbb{E}_S[R(A(S)) - R(w^*)] = O\bigg( n^{-\frac{3}{4}}\|w^*\|\sqrt{R(w^*)} + n^{-\frac{3}{2}}\|w^*\|^2 \bigg) = O(n^{-1}\|w^*\|^2).
\end{aligned}
\end{equation*}
\end{remark6proof}

\subsection{B.5~~~Proofs of Corollaries 1-3}
\begin{corollary1proof}
To apply Theorem 2, we just need to verify the loss function in triplet metric learning satisfies convex, $\alpha$-smooth and $L$-Lipschitz.

{\bf{Convexity.}} For all $w$, $z^+, \tilde{z}^+, z^-$,
\begin{equation}
\label{Equation 30}
    \nabla^2 \ell_{\phi}(w; z^+, \tilde{z}^+, z^-) = \Big( (x^+ - \tilde{x}^+) (x^+ - \tilde{x}^+)^\top - (x^+ - x^-) (x^+ - x^-)^\top \Big)^2 \phi^{\prime \prime} \big(h_w(x^+, \tilde{x}^+) - h_w(x^+, x^-) + \zeta\big).
\end{equation}
It is easy to verify that $\nabla^2 \ell_{\phi}(w; z^+, \tilde{z}^+, z^-) \geq 0$ since the second derivative $\phi^{\prime \prime}$ of logistic function always is non-negative.

{\bf{Lipschitz continuity.}} For all $w$, $w^\prime$, $z^+$, $\tilde{z}^+$, $z^-$,
\begin{equation}
\label{Equation 31}
\begin{aligned}
    & \Big| \ell_{\phi}(w; z^+, \tilde{z}^+, z^-) - \ell_{\phi}(w^{\prime}; z^+, \tilde{z}^+, z^-)\Big|\\
    = & \Big| \mathrm{log}\big( 1 + \mathrm{exp}(h_w(x^+, x^-) - h_w(x^+, \tilde{x}^+) - \zeta) \big) - \mathrm{log}\big( 1 + \mathrm{exp}(h_{w^{\prime}}(x^+, x^-) - h_{w^{\prime}}(x^+, \tilde{x}^+) - \zeta) \big) \Big|\\
    \leq & \Big| h_w(x^+, x^-) - h_w(x^+, \tilde{x}^+) - \zeta - h_{w^{\prime}}(x^+, x^-) + h_{w^{\prime}}(x^+, \tilde{x}^+) + \zeta \Big|\\
    = & \Big| \big\langle w-w^{\prime}, (x^+ - x^-)(x^+ - x^-)^\top \big\rangle - \big\langle w-w^{\prime}, (x^+ - \tilde{x}^+)(x^+ - \tilde{x}^+)^\top \big\rangle \Big|\\
    \leq & \Big| \big\langle w-w^{\prime}, (x^+ - x^-)(x^+ - x^-)^\top \big\rangle \Big| + \Big| \big\langle w-w^{\prime}, (x^+ - \tilde{x}^+)(x^+ - \tilde{x}^+)^\top \big\rangle \Big|\\
    \leq & 8B^2 \|w - w^{\prime}\|,
\end{aligned}
\end{equation}
where the first inequality is due to the 1-smoothness of the logistic function, and the last two inequalities are obtained with the Cauchy-Schwartz inequality and the boundedness assumption of sample space.

{\bf{Smoothness.}} With the similar fashion as above, we deduce that
\begin{equation*}
\begin{aligned}
    & \Big|\Big| \nabla \ell_{\phi}(w; z^+, \tilde{z}^+, z^-) - \nabla \ell_{\phi}(w^\prime; z^+, \tilde{z}^+, z^-) \Big|\Big|_2\\
    \leq & \Big|\Big| (x^+ - \tilde{x}^+)(x^+ - \tilde{x}^+)^\top - (x^+ - x^-)(x^+ - x^-)^\top \Big|\Big|_2 \cdot\\
    & \Big| \phi^\prime \big( h_w(x^+, \tilde{x}^+) - h_w(x^+, x^-) + \zeta \big) - \phi^\prime \big( h_{w^\prime}(x^+, \tilde{x}^+) - h_{w^\prime}(x^+, x^-) + \zeta \big) \Big|\\
    \leq & \Bigg( \Big|\Big| (x^+ - \tilde{x}^+)(x^+ - \tilde{x}^+)^\top \Big|\Big|_2 + \Big|\Big| (x^+ - x^-)(x^+ - x^-)^\top \Big|\Big|_2 \Bigg) \Big| \phi^\prime \big( h_w(x^+, \tilde{x}^+) - h_w(x^+, x^-) + \zeta \big)\\
    & - \phi^\prime \big( h_{w^\prime}(x^+, \tilde{x}^+) - h_{w^\prime}(x^+, x^-) + \zeta \big) \Big|\\
    = & 8B^2 \Big| \phi^\prime \big( h_w(x^+, \tilde{x}^+) - h_w(x^+, x^-) + \zeta \big) - \phi^\prime \big( h_{w^\prime}(x^+, \tilde{x}^+) - h_{w^\prime}(x^+, x^-) + \zeta \big) \Big|\\
    \leq & 8B^2 \Big| h_w(x^+, x^-) - h_w(x^+, \tilde{x}^+) - \zeta - h_{w^{\prime}}(x^+, x^-) + h_{w^{\prime}}(x^+, \tilde{x}^+) + \zeta \Big|\\
    \leq & 64B^4 \|w - w^\prime\|.
\end{aligned}
\end{equation*}
Therefore, the desired result is obtained by Theorem 2, where loss function satisfies the convexity, $8B^2$-Lipschitz continuity and $64B^4$-smoothness.
\end{corollary1proof}

\begin{corollary2proof}
To apply Theorem 3, we need to prove the $\sigma$-strong convexity of $F_S$ and $L$-Lipschitz continuity of $\ell$.
They can be verified  by \eqref{regularization} and  \eqref{Equation 31}.
Thus, we get the stated result from Theorem 3.

\end{corollary2proof}

\begin{corollary3proof}
From the proofs of Corollaries 1 and 2, we know that the $\sigma$-strong convexity of $F_S$ and the $\alpha$-smoothness of $\ell$. Hence, we can directly apply Theorem 4 to get the desired result.
\end{corollary3proof}

\section{C. ~~Review on the Definitions of  Algorithmic Stability} 
In machine learning literature, there are various definitions of algorithmic stability for pointwise/pairwise learning, including hypothesis stability \cite{DBLP:journals/jmlr/BousquetE02, DBLP:journals/jmlr/ElisseeffEP05}, error stability \cite{DBLP:journals/jmlr/BousquetE02, DBLP:journals/jmlr/Shalev-ShwartzSSS10}, uniform stability \cite{DBLP:journals/jmlr/AgarwalN09, DBLP:conf/icml/HardtRS16,  DBLP:conf/nips/FosterGKLMS19, DBLP:conf/nips/FeldmanV18, DBLP:conf/colt/FeldmanV19, DBLP:conf/colt/BousquetKZ20, DBLP:conf/nips/KlochkovZ21, DBLP:conf/aaai/SunLW21, DBLP:journals/corr/ShenYYY19, DBLP:conf/colt/Maurer17, DBLP:conf/ijcai/GaoZ13, DBLP:conf/nips/JinWZ09, DBLP:conf/aistats/WangZLSP19, DBLP:journals/corr/ChenJY2018}, uniform augment stability \cite{DBLP:conf/icml/LiuLNT17, DBLP:conf/nips/BassilyFGT20, DBLP:conf/nips/XingSC21, DBLP:conf/nips/YangLWYY21, DBLP:conf/icml/LeiYYY21}, on-average stability \cite{DBLP:conf/nips/LeiLK20, DBLP:conf/iclr/LeiY21, DBLP:conf/icml/KuzborskijL18}, on-average augment stability \cite{DBLP:conf/icml/LeiY20, DBLP:conf/nips/LeiLY21} and locally elastic stability \cite{DBLP:conf/icml/DengHS21}. 

It is well known the above stability definitions for pointwise/pairwise learning is the building block for triplet algorithmic stability. 
Therefore, to better understand our Definitions 1 and 2 for triplet learning, we summarize the previous definitions of algorithmic stability in Table \ref{Notions of various stability}, where their properties are illustrated from the aspects of the measure of perturbation (\emph{Loss $\ell$ Vs. Model} $A(S)$) and the dependence of stability parameter (\emph{data dependence Vs. data independence}). 
Here, let $S=\{z_1,...,z_n\}, \bar{S}=\{\bar{z}_1,...,\bar{z}_n\}$ be drawn independently from the same data generating distribution, and let $S_i=\{z_1, ..., z_{i-1}, \bar{z}_i, z_{i+1}, ..., z_n\}$. 
\begin{table}[h]
    \centering
   
    \renewcommand\arraystretch{1.5}
    \begin{tabular}{l|lccc}
        \hline
        Stability & Definitions & Loss & Model & \thead{Data \\ dependence}\\
        \hline
        Hypothesis stability & $\mathbb{E}_{S, \bar{S}, z}[|\ell(A(S), z) - \ell(A(S_i), z)|] \leq \gamma, \forall i\in [n]$ & $\surd$ & $\times$ & $\times$\\
        Error stability & $|\mathbb{E}_z[\ell(A(S), z) - \ell(A(S_i), z)]| \leq \gamma, \forall S, \bar{S}\in \mathcal{Z}^n, \forall i\in [n]$ & $\surd$ & $\times$ & $\times$\\
        Uniform stability & $\|\ell(A(S), \cdot) - \ell(A(S_i), \cdot)\|_\infty \leq \gamma, \forall S, \bar{S}\in \mathcal{Z}^n, \forall i\in [n]$ & $\surd$ & $\times$ & $\times$\\
        Uniform augment stability & $\|A(S) - A(S_i)\|_\infty \leq \gamma, \forall S, \bar{S}\in \mathcal{Z}^n, \forall i\in [n]$ & $\times$ & $\surd$ & $\times$\\
        On-average stability & $\frac{1}{n} \sum\limits_{i=1}^n \mathbb{E}_{S, \bar{S}}[\ell(A(S), z_i) - \ell(A(S_i), z_i)] \leq \gamma, \forall S, \bar{S}\in \mathcal{Z}^n$ & $\surd$ & $\times$ & $\times$\\
        On-average augment stability & $\frac{1}{n} \sum\limits_{i=1}^n \mathbb{E}_{S, \bar{S}} [\|A(S) - A(S_i)\|]\leq \gamma, \forall S, \bar{S}\in \mathcal{Z}^n$ & $\times$ & $\surd$ & $\times$\\
        Locally elastic stability & $|\ell(A(S), z) - \ell(A(S_i), z)| \leq \gamma(z_i, z), \forall S, \bar{S}\in \mathcal{Z}^n, \forall i\in [n]$ & $\surd$ & $\times$ & $\surd$\\
        \hline
   
    \end{tabular}
          \caption{Summary of definitions and properties of algorithmic stability ($\surd$-has such a property; $\times$-hasn't such a property)}
    \label{Notions of various stability}
\end{table}

From Table \ref{Notions of various stability}, we know that the uniform stability (uniform augment stability) is stronger than the on-average stability (on-average augment stability). 
As demonstrated in \citet{DBLP:conf/icml/DengHS21}, the locally elastic stability is a more fine-grained stability than the others due to its data dependence of stability parameter $\gamma$. 

From the other side, the uniform (on-average) augment stability implies the uniform (on-average) stability when the loss function satisfies $L$-Lipschitz continuous. 
That is to say, the model-based stability characterization is stronger than the loss-based stability measure usually. 
In addition, the average augment stability has various versions when employing different norms.

Table \ref{Notions of various stability} just shows the definitions in the case of changes of one point. 
It has been extended to the setting of changes of two points \cite{DBLP:conf/nips/LeiLK20, DBLP:conf/nips/LeiLY21} and the setting of multitask learning \cite{DBLP:conf/aistats/WangZLSP19, DBLP:conf/aaai/Zhang15}. 
These developments pave the way to motivate our Definitions 1 and 2.

\section{D.~~ Discussion}
There are still many interesting research topics related to the current work, 
which we will discuss below:

{\bf{Non-convex loss functions:}} As shown in Tables 1-2, our stability-based generalization analysis requires the convexity of triplet loss function. 
However, there also involve various non-convex loss functions in some triplet learning algorithms. 
Hence, it is important to investigate the generalization and stability for general non-convex triplet losses. 
In addition, some restrictions on the non-convex loss functions may be necessary, e.g., the Polyak-${\L}$ojasiewicz (PL) condition and the quadratic growth (QG) condition \cite{DBLP:conf/icml/CharlesP18}. 

{\bf{Distribution Shift:}}  To the best of our knowledge, the existing stability-based generalization bounds are all under the assumption that the testing data and the training data are drawn independently from an identity distribution. 
Motivated by the widespread practical deployment of learning algorithms, there often faces complicated data environment where the unknown test distribution potentially differs from the training distribution, i.e., distribution shift \cite{DBLP:conf/ijcai/WangLLOQ21, DBLP:journals/corr/ShenLHZXYC2021, DBLP:conf/colt/AgarwalZ22}. 
For a wide range of models and distribution shifts, \citet{DBLP:conf/icml/MillerTRSKSLCS21} shows the strong empirical correlation between out-of-distribution performance and in-distribution performance. 
In particular, there is rapid theoretical progress on the generalization guarantees under distribution shift by leveraging the expansion assumption \cite{DBLP:conf/nips/YeXCLLW21, DBLP:conf/icml/CaiGLL21}, the uniformly convergence analysis \cite{DBLP:conf/colt/AgarwalZ22}, and the operator approximation \cite{Appl.Comput.Harmon.Anal/GizewskiMMNPPSZ21}. Therefore, it is natural and crucial to further investigate the stability-based generalization theory  \cite{DBLP:conf/icml/HardtRS16, DBLP:conf/icml/CharlesP18, DBLP:conf/icml/LeiY20} for the distribution shift setting. 

\end{document}